\documentclass[lettersize,journal]{IEEEtran}
\usepackage{amsmath,amsfonts}
\usepackage{algorithmic}
\usepackage{algorithm}
\usepackage{array}
\usepackage[caption=false,font=normalsize,labelfont=sf,textfont=sf]{subfig}
\usepackage{textcomp}
\usepackage{stfloats}
\usepackage{url}
\usepackage{verbatim}
\usepackage{graphicx}
\usepackage{enumitem}
\usepackage{float}
\usepackage{multirow}
\usepackage{cite}
\begin{document}

\title{Efficient Video Segmentation Models with Per-frame Inference}

\author{Yifan Liu$^{1}$         \and
        Chunhua Shen$^{2}$ \and
        Changqian Yu$^{3}$ \and
        Jingdong Wang$^4$%
\IEEEcompsocitemizethanks
{\IEEEcompsocthanksitem Y. Liu is with
The University of Adelaide, SA 5005, Australia. 
\protect\\
E-mail: yifan.liu04@adelaide.edu.au
\IEEEcompsocthanksitem C. Shen is with Zhejiang University, Hangzhou, China.
\IEEEcompsocthanksitem C. Yu is with MeiTuan, Beijing, China.
\IEEEcompsocthanksitem J. Wang is with AI Group, Baidu, Beijing, China.\protect \\
}
}%

\maketitle

\begin{abstract}
Most existing real-time deep models trained with each frame independently may produce inconsistent results 
across the temporal axis 
when tested on
a video sequence. 
A few methods take %
the correlations in the video sequence into account, \textit{e.g.}, by propagating the results to the neighboring frames using optical flow, or extracting frame representations using multi-frame information,
which may lead to inaccurate results or unbalanced latency. { In this work, we focus on improving the temporal consistency without introducing computation overhead in inference.
To this end, we perform inference at each frame. Temporal consistency is achieved 
by learning from %
video frames with extra constraints during the training phase. 
introduced for inference.
We propose several %
techniques to learn from the video sequence, including a temporal 
consistency 
loss and online/offline knowledge distillation methods.} 
On the task of semantic video segmentation, 
weighing among accuracy, temporal sm\-oo\-th\-ne\-ss, and efficiency, our proposed method outperforms 
keyframe based methods and 
a few 
baseline methods that 
are trained with each frame  independently, 
on  datasets including Cityscapes, Camvid { and 300VW-Mask}. We further apply our training method to { video instance segmentation on YouTubeVIS and develop an application of portrait matting} in video sequences, by segmenting 
temporally  consistent
instance-level trimaps across frames.
Experiments show superior qualitative and quantitative results.  
Code is available at: \url{https://git.io/vidseg}.

\end{abstract}

\section{Introduction}

\label{intro}

\IEEEPARstart{B}{enefiting} from the powerful convolutional neural networks (CNNs), some recent methods~\cite{chen2018deeplab,wang2020solov2,tian2019fcos} have achieved outstanding performance 
for high-level scene understanding tasks %
such as 
semantic segmentation, instance segmentation, and object detection. Typically, these techniques address the problem of scene analysis from a single image. Applying them to each frame independently may produce 
temporally 
inconsistent results when tested on a video sequence.
One cause is 
the lack of high-quality labeled video data.  
For most popular benchmarks~\cite{Cordts2016Cityscapes,brostow2008segmentation}, 
there is only one labeled frame every a few frames. When training on the labeled images, the per-frame models\footnote{We denote the model 
that 
takes single frames as inputs when testing on the video sequence as per-frame models in this paper.} %
can 
be sensitive to small jitters, and then produce inconsistent results. 

{ Another reason may be the neglecting of the correlations among frames. When training on each image independently, the models do not care about the consistency between similar inputs, but only tend to minimize the difference between the prediction and the ground truth labels on a single frame, leading to inconsistent results.
\begin{figure}[htbp]
    \centering 
    \includegraphics[width=0.5\textwidth]{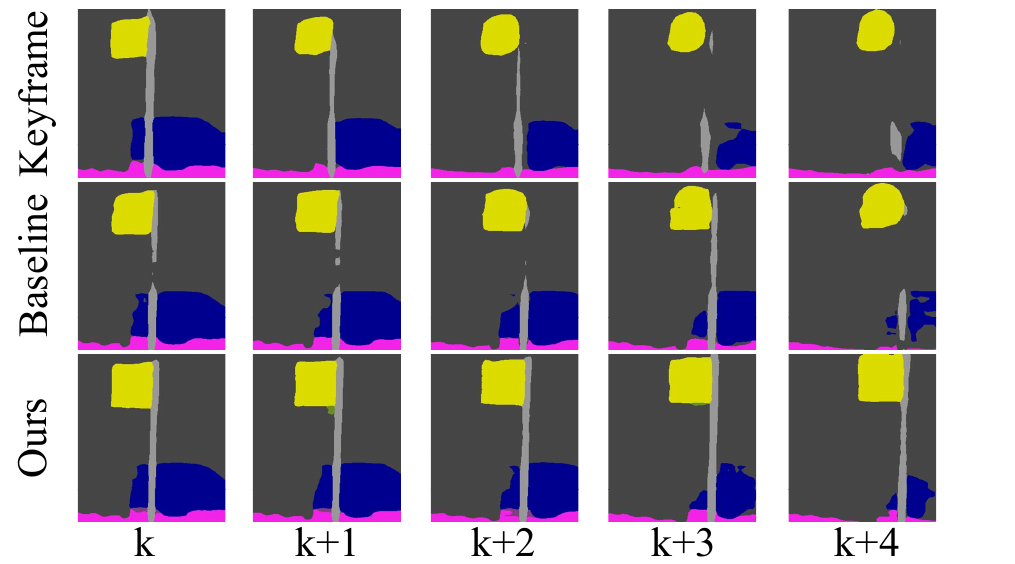}
\     \caption{We take semantic segmentation as an example. Visualization results on consecutive frames: \textit{Keyframe}: Accel18~\cite{jain2019accel} propagates and fuses the results from the keyframe ($k$) to 
non-key
frames ($k+1,\dots$), which %
may lead to
unsatisfactory 
results on %
non-key
frames. \textit{Baseline}: PSPNet18~\cite{zhao2017pyramid} trains the model on single frames. 
Inference on single frames separately  can
produce temporally inconsistent results. 
\textit{Ours}: training the model with the correlations among frames and inferring %
on single frames separately %
lead to
a good balance between high-quality 
results and inference time.
}
    \label{Fig.first}
\end{figure}

To achieve temporally consistent predictions, one needs to take  the smoothness constraints into account during either training/inference, or both.} Previous methods perform
post-processing among frames~\cite{liu2017surveillance,zhang2019exploiting}, or 
employ 
extra modules to use 
multi-frame 
information during training and inference~\cite{gadde2017semantic}. The high computational cost makes it difficult for mobile applications. 
To speed up processing, 
keyframes approaches %
avoid processing of each frame, and 
propagate  the outputs or the feature maps to other frames (%
non-key
frames)
using 
optical flows \cite{zhu2017deep,zhu2018towards,xu2018dynamic}. %
Keyframe based methods 
indeed accelerate inference. 
However, it requires different inference time for keyframes and 
non-key
frames, 
leading to an unbalanced latency, thus being not friendly for real-time processing. 
Another drawback is that the accuracy cannot be guaranteed for each frame due to the cumulative warping error, for example, the first row in Figure~\ref{Fig.first}.

Recently, real-time models on %
images~\cite{Sandler2018MobileNetV2IR,yu2020bisenet,mehta2019espnetv2} have draw much attention. Clearly, applying compact networks to each frame of %
a  video 
sequence independently may alleviate the latency and enable 
real-time execution. However, as discussed above, directly training the model on each frame independently 
often
produces 
temporally 
inconsistent results on  videos %
as shown in the second row of Figure~\ref{Fig.first}. 

It is important to %
achieve a good balance between 
accuracy, temporal consistency, and efficiency when designing algorithms for video scene understanding. As discussed above,  employing per-frame inference can ease the problem of latency and achieve stable prediction time. %
{  In this work, we propose a semi-supervised 
learning scheme to train  per-frame  models  on video sequences to obtain improved accuracy and temporal consistency. Several alternative techniques are employed, including a temporal consistency loss considering the correlation among frames and online/offline knowledge distillation.
}

A motion guided \textit{temporal consistency loss} is employed with the motivation of assigning a consistent label to the same pixel along 
the time axis. Occlusions are masked by self-supervision from RGB images, making 
the model more robust to the small jitters.

{
Knowledge distillation methods are widely used in image recognition tasks~\cite{liu2019structured,he2019knowledge,Li2017MimickingVE}, and achieve great success. To further improve the performance of the per-frame models, we propose 
an 
online \textit{temporal consistency knowledge distillation} strategy, which learns the correlation among frames from %
a 
large teacher network. Besides, an offline knowledge distillation scheme, generating pseudo labels for unlabeled consecutive frames, is employed. The online and 
offline knowledge distillation methods can boost not only the accuracy, but also the smoothness, verifying our hypothesis that learning from video frames can ease the problem of temporally inconsistent predictions.
}

\begin{figure*}[htbp]
    \centering 
    \subfloat[Temporal consistency vs. inference speed]{\includegraphics[width=0.493\textwidth]{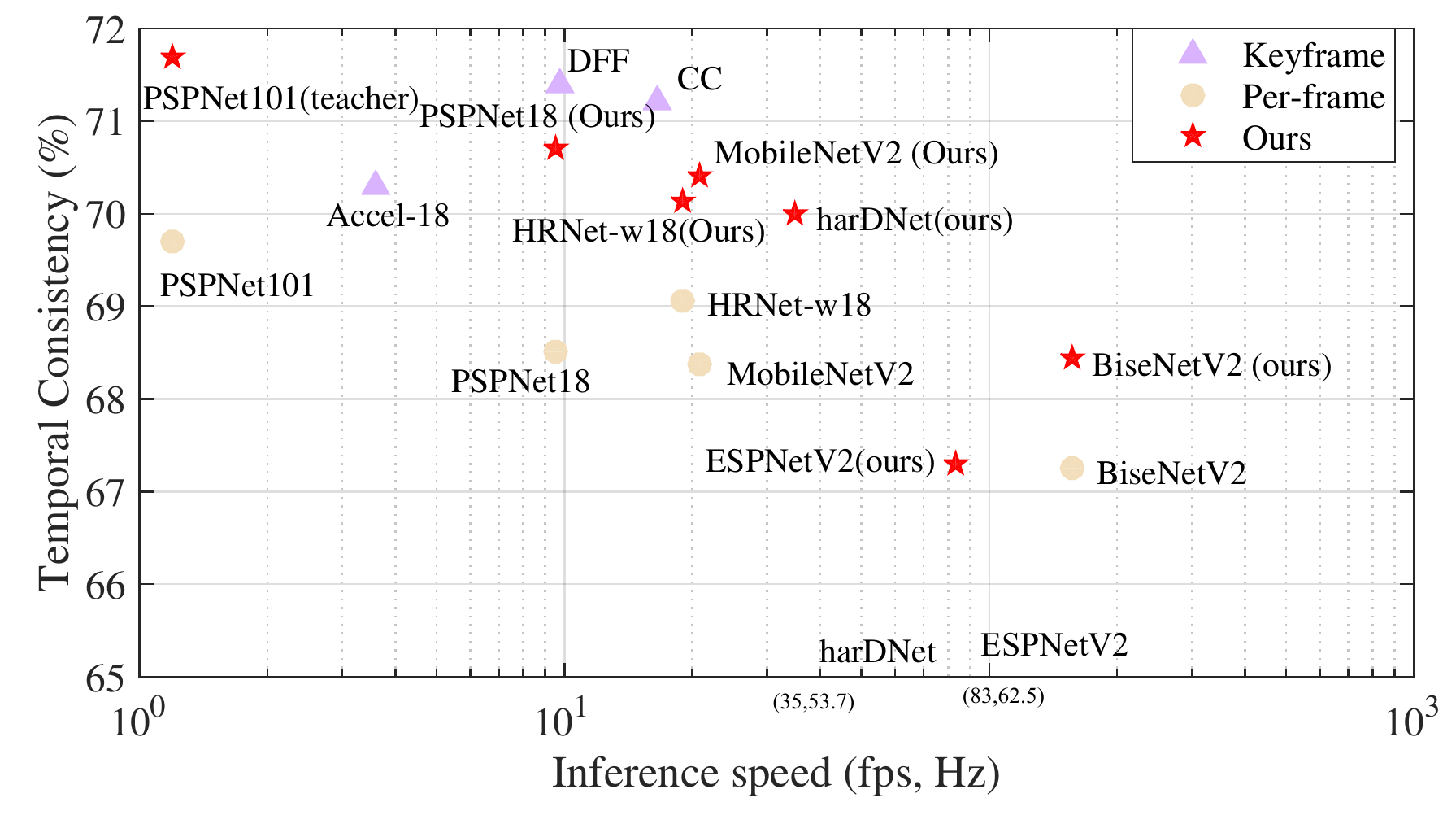}}
\subfloat[Accuracy vs. inference speed]{\includegraphics[width=0.485\textwidth]{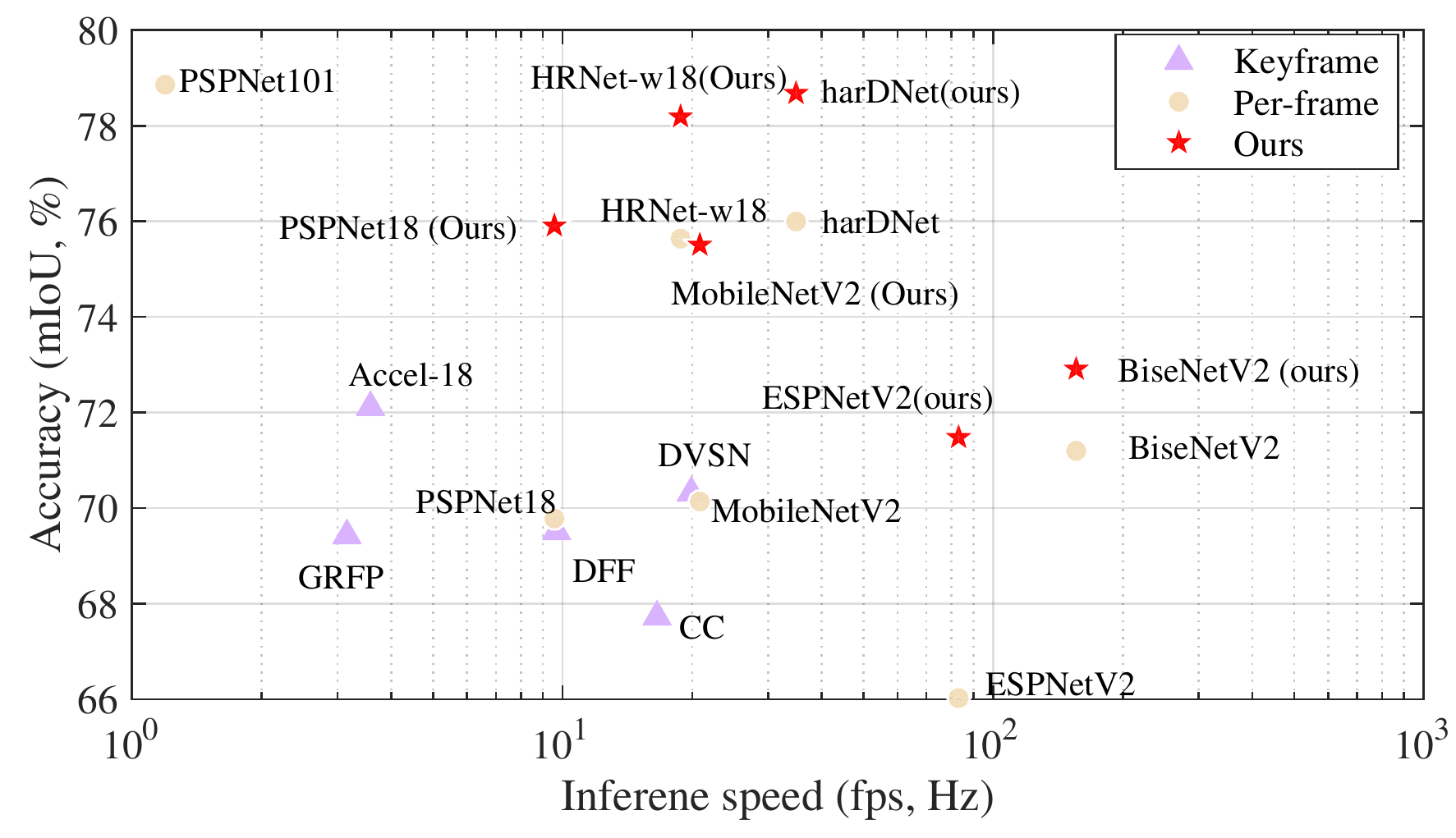}\label{Fig.tc-t}}

\caption{Comparing our %
proposed 
per-frame models with a few %
keyframe based methods: Accel~\cite{jain2019accel}, DVSN~\cite{xu2018dynamic}, DFF~\cite{zhu2017deep} and  CC~\cite{shelhamer2016clockwork}. The inference speed of harDNet~\cite{chao2019hardnet}, ESPNetV2~\cite{mehta2019espnetv2} and BiseNetV2~\cite{yu2020bisenet} are reported in their papers. The inference speed of other methods is evaluated on a single GTX 1080Ti.}
    \label{Fig.acc-t}
\end{figure*}

{We first validate the effectiveness of proposed training methods with efficient semantic video segmentation on %
benchmark datasets.
The performance is first evaluated on two self-driving datasets, Cityscapes~\cite{Cordts2016Cityscapes} and Camvid~\cite{brostow2008segmentation}. Results of face mask segmentation on 300VW-mask~\cite{wang2019face} are also reported. The temporal consistency as well as the segmentation accuracy are boosted compared to the per-frame models, without extra computation cost during inference as shown in Figure~\ref{Fig.acc-t}. Weighing among accuracy, temporal consistency, and efficiency, the improved per-frame models can achieve promising performance.

Moreover, we apply our training methods to the recent instance segmentation model \cite{wang2020solov2}, and boost its performance on YouTube-VIS \cite{Yang_2019_ICCV}. With the per-frame instance segmentation network, we propose an 
application of predicting instance-level trimaps (foreground, background, unknown), which can be used to generate instance-level portrait matting. Training with pseudo labels and the motion loss does not depend on the ground-truth labels. Thus, the proposed training scheme can be easily applied to the videos in the wild.  Qualitative and quantitative results on images and videos demonstrate the proposed training methods are helpful for transfer learning %
and can improve the temporal consistency of the per-frame models in %
a few 
applications.
}

We summarize our main contributions as follows.
\begin{itemize}
    \setlength{\itemsep}{0pt}
    \setlength{\parskip}{0pt}
    \setlength{\parsep}{0pt}
    \item We consider the correlations among frames during training and independently process each frame in the inference process for video segmentation, thus improving temporal consistency without introducing computation overhead in inference. 
    
    \item {We employ %
    a few 
    techniques, including a temporal consistency loss and online/offline knowledge distillation schemes to %
    effectively train 
    the per-frame models, taking both accuracy and temporal consistency into consideration. }
    \item When applying to semantic video segmentation, the compact models outperform previous state-of-the-art methods weighing among accuracy, temporal consistency, and inference speed on datasets including Cityscapes, Camvid and 300VW-Mask.
    \item We extend the proposed training methods to video instance segmentation, and apply it to portraits matting in the wild.
\end{itemize}
A preliminary version of this work was published in~\cite{liu2020efficient}. We have extended our conference
version as follows. {First, we simplify the original training scheme by removing the online single frame distillation. %
As a result, memory footprint in training is reduced. 
Second, 
we employ off-line distillation with test-time augmentation to generate pseudo labels on video sequences, and significantly 
improve 
the accuracy and the temporal consistency.
Last, 
we %
 include more experiments on 300VM-Mask and apply 
the method to 
video instance segmentation on YouTube-VIS. We propose an application of video portraits matting in the wild by predicting instance-level trimaps. }

\section{Related Work}

\noindent\textbf{Semantic Video Segmentation.} Semantic video segmentation requires dense labeling for all pixels in each frame of a video sequence into a few semantic categories. 
Previous work can be summarized into two streams.
The first one focuses on improving the accuracy by exploiting the temporal relations and the unlabelled data in the video sequence. A gated recurrent unit is employed to propagate semantic labels to unlabeled frames in \cite{nilsson2018semantic}. Other works such as
NetWarp~\cite{gadde2017semantic}, STFCN~\cite{fayyaz2016stfcn}, and SVP~\cite{liu2017surveillance} %
employ optical-flow or recurrent units to fuse the results of several frames during inferring to improve the segmentation accuracy. 

The second line of works
pay attention to reduce the computation cost by re-using the feature maps in the neighboring frames. ClockNet~\cite{shelhamer2016clockwork} proposes to copy the feature map to the next frame directly,
thus reducing the computational cost.
DFF~\cite{zhu2017deep} employs the optical flow to warp the feature map between the keyframe and %
non-key 
frames.  Xu \textit{et al.} 
\cite{xu2018dynamic} further propose to use an adaptive keyframe selection policy while Zhu \textit{et al.} \cite{zhu2018towards}
observe 
that propagating partial regions in the feature map can %
achieve 
better performance. Li \textit{et al.}~\cite{li2018low} propose a low-latency video segmentation network by optimizing both the keyframe selection and the adaptive feature propagation. Accel~\cite{jain2019accel} proposes a network fusion policy to use a large model to predict the keyframe and use a compact one in
non-key
frames.
Keyframe based methods require different inference time and may produce different quantity results between keyframes and 
other 
frames.

\noindent\textbf{Video/image Instance Segmentation.}
Image instance segmentation is a fundamental problem in computer vision, 
which predicts each pixel a category label as well as an instance label, \textit{a.k.a.}\ pixel-level object detection.
Typical methods often follow the pipeline of  `detect-then-segment'  \cite{he2017mask,kirillov2019pointrend}, which need to predict 
a bounding box for each instance first and then perform binary segmentation within each box.
To avoid heavy computation, often  
the resolution of the estimated masks is of low resolution, \textit{e.g.},  
$28 \times 28$ in Mask RCNN \cite{he2017mask}.
Recently, several
approaches~\cite{wang2019solo, tian2020conditional,wang2020solov2} solve 
instance segmentation without resorting to bounding box detection. { MaskTrack R-CNN \cite{Yang_2019_ICCV} first proposes the task and a new dataset for video instance segmentation. They extend the Mask-RCNN to video instance segmentation tasks by adding a new tracking branch. In this work, we aim at using simple training schemes to perform video instance segmentation with recent state-of-the-art one-stage image instance segmentation networks~\cite{wang2019solo,wang2020solov2}.}
 
 {
\noindent\textbf{Trimaps Prediction.} Trimaps are usually employed as the input of a matting system.  It is challenging to perform video matting as one needs to make sure the temporal consistency across frames. 
Trimaps are often generated by a binary semantic segmentation method 
and/or interactive user inputs. 
Current research focuses on generating consistent trimaps across video frames. 
Flow-based methods~\cite{lee2010temporally,bai2011towards} 
propagate the generated trimaps using motion information. Zou \emph{et al.} ~\cite{zou2019unsupervised} propose an unsupervised video matting method with sparse and low-rank representations. Different from previous work, we apply our per-frame model 
to 
a new application of instance portraits matting by instance-level trimaps prediction. }

\noindent\textbf{Temporal Consistency.}
Applying image processing algorithms to each frame of a video may lead to
temporally 
inconsistent results. The temporal consistency problem has draw much attention in low-level and mid-level applications, such as 
colorization~\cite{levin2004colorization}, style transfer~\cite{gupta2017characterizing}, and video depth estimation~\cite{bian2019unsupervised,bian2020unsupervised} and task 
agnostic 
approaches~\cite{lai2018learning,yao2017occlusion}.  
Temporal consistency is also essential in semantic video segmentation.
Miksik \textit{et al.}~\cite{miksik2013efficient} employ a post-processing method that learns a similarity function between pixels of consecutive frames to propagate predictions across time. Nilsson and Sminchiesescu~\cite{nilsson2018semantic} insert the optical flow estimation network into the forward pass and employ a recurrent unit to make use of neighbouring predictions. Our method is more efficient than
theirs 
as we employ  per-frame inference.
The warped previous predictions work as a constraint
\textit{only} during training.

\noindent\textbf{Knowledge Distillation.}
The effectiveness of knowledge distillation has been verified in classification~\cite{hinton2015distilling,romero2014fitnets,Zagoruyko2016PayingMA}. The output of the large teacher net, including the final logits and the intermediate feature maps, are %
used 
as soft targets to supervise the compact student net. 
Previous knowledge distillation methods in semantic segmentation~\cite{he2019knowledge,liu2019structured} design distillation 
strategies 
only for improving the segmentation accuracy.
To our knowledge, to date no distillation methods are %
employed 
to improve 
temporal consistency. In this work, we focus on encoding the motion information into the distillation 
terms
to make the segmentation networks %
work better 
for the semantic video segmentation tasks.

\noindent\textbf{Semi-supervised learning.} { Pseudo labels are widely used in semi-supervised learning. Mittal \emph{et al.}~\cite{mittal2019semi} 
propose an approach for semi-supervised semantic segmentation that learns from
a small amount of pixel-wise annotated samples while exploiting additional annotation-free images. 
Ouali \emph{et al.}~\cite{ouali2020semi} use pseudo labels to train an auxiliary branch. These works %
assume that one only has 
limited labeled data, and the generated pseudo labels are of low quality. Thus, they make extra efforts to take advantage of the pseudo labels. Recently, directly training the network with high-quality pseudo labels has gained much attention~\cite{zhu2019improving,chen2020naive}. Zhu \textit{et al.} ~\cite{zhu2019improving} propose to use a motion estimation network to propagate labels to unlabeled frames as data augmentation and achieve state-of-the-art performance in terms of  segmentation accuracy.  Chen~\textit{et al.} ~\cite{chen2020naive} propose an iterative training scheme to train the student network with pseudo labels. These methods %
achieve significant performance.
DVRL~\cite{wang2020dynamic} employ pseudo labels to evaluate the similarity between frames as the reward of reinforcement learning. Thus, they can select key-frames based on the rewards.

Different from previous work, in this paper, we focus on training efficient per-frame models on video sequences to improve temporal consistency and accuracy. Training with the pseudo labels on consecutive frames can improve the accuracy as well as the smoothness. 
}

\section{Our Approach}

\begin{figure*}[htbp]
    \centering 
    \includegraphics[width=1.0\textwidth]{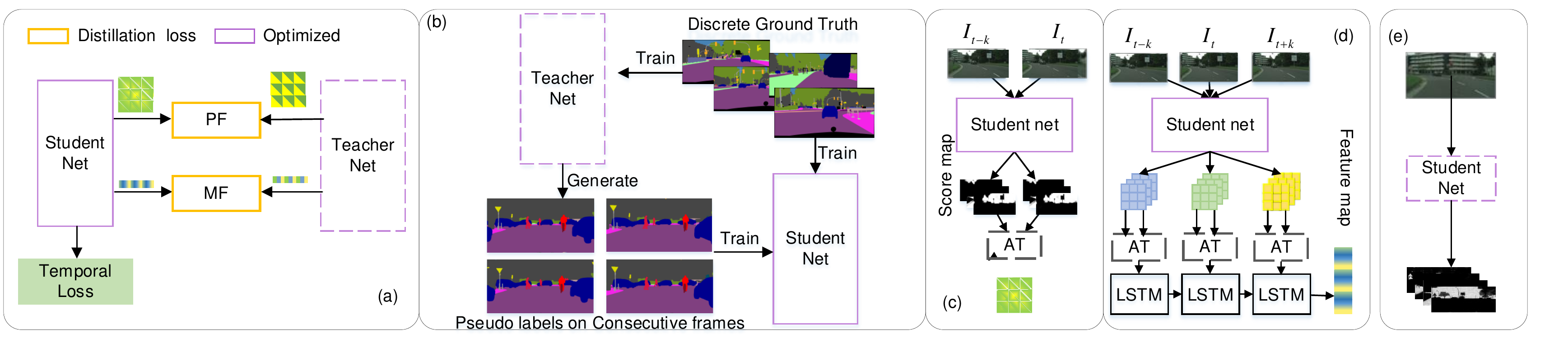}
\     \caption{{ (a) \textbf{Overall of the training loss terms}: The distillation (c and d) loss and the temporal loss (Figure~\ref{fig:tl}) are considered during training. (b) \textbf{Overall of the training pipeline}: The teacher net is trained on the ground truth labels, and used to generate pseudo labels on consecutive frames.  (c) \textbf{Pair-wise frame dependency (PF)}: encode the motion relations between two frames. (d) \textbf{multi-frame dependency (MF)}: extract the correlations of the intermediate feature maps among multi-frames.  (e) \textbf{The inference process}. All the proposed methods are only applied during training. We can improve the temporal consistency as well as the segmentation accuracy without any extra parameters or post-processing during inference.}}
    \label{fig:overall}
\end{figure*}

{ Figure~\ref{fig:overall} 
summarizes 
our main contributions. 
A few 
training techniques are employed to improve the accuracy and the temporal consistency of the per-frame models, including the offline distillation (Figure~\ref{fig:overall} (b)), the online distillation (Figure~\ref{fig:overall} (c) and (d)) and the temporal loss (Figure~\ref{fig:tl}). In this section, taking the semantic video segmentation as an example, we %
first describe the overall training scheme in Section~\ref{sec:PL}. Then, we introduce the details of the training loss, including the online distillation loss in Section~\ref{sec:tckd} and  the temporal loss in Section~\ref{sec:TL}. Finally, we extend the training scheme to video instance segmentation. }

\subsection{Self-training on Video Sequences}
\label{sec:PL}

Without supervision of temporal consistency during training and inference, 
the network is sensitive to the small jitters in the input frame. Here, we 
propose a training scheme which considers the temporal consistency and makes use of unlabeled video frames. The details are shown in Algorithm~\ref{alg:semi}.

Inspired by~\cite{liu2019structured}, we build a distillation mechanism to effectively train the compact student net $\mathsf{S}$ by making use of the cumbersome teacher net $\mathsf{T}$. The teacher net $\mathsf{T}$ is already well trained with  the segmentation loss  $\ell_{ce}$ and the temporal loss $\ell_{tl}$ to achieve a high temporal consistency as well as the segmentation accuracy. Then, the pseudo labels  on consecutive frames are generated by $\mathsf{T}$ with test-time data augmentation. We build a new training set by combining pseudo labels from the video frames with ground-truth labels from the images. 
The student net $\mathsf{S}$ is trained on the combined dataset with three losses: 
the conventional cross-entropy loss $\ell_{ce}$, temporal loss $\ell_{tl}$, and the distillation loss $\ell_{dis}$:
\begin{equation}
        \ell=\sum_{t=1}^{T}\ell_{ce}^{(t)}+\lambda (\sum_{i=1}^{T-1}\ell_{tl}(\mathbf{Q}_t,\mathbf{Q}_{t+1})
        +\sum_{i=1}^{T-1}\ell_{dis}),
 \label{eq:all}
\end{equation}

\noindent where $T$ is the number of all the frames in one training sequence including frames with ground-truth labels and pseudo labels. $\mathbf{Q}_t$ and $\mathbf{Q}_{t+1}$ is the segmentation map for frame $t$ and frame $t+1$. The loss weight for all regularization terms $\lambda$ is set to $0.1$. Different from previous single frame distillation methods~\cite{liu2019structured},
$\ell_{dis}$ contains two new distillation 
strategies, which 
are designed to transfer the temporal consistency from $\mathsf{T}$ to $\mathsf{S}$: pair-wise-frames dependency (PF) and multi-frame dependency (MF).

\begin{algorithm}[!t]
\centering
\caption{
Temporally 
Consistent Semi-supervised Learning for  Video Segmentation}
\label{alg:semi}
\begin{algorithmic}
\STATE \textbf{Labeled data}: $n$ pairs of image $x_i$ and corresponding human annotation $y_i$
\STATE \textbf{Unlabeled data}: $m$ images collected from multiple video sequences with no human annotations $\{\widetilde{x}_1, \widetilde{x}_2, ..., \widetilde{x}_m\}$.
\STATE \textbf{Step 1}: Train a Teacher network $\theta_{t}$ on labeled images by minimizing $\ell_t$.
  \IF{sampled images have consecutive frames}
  \STATE
  $\ell_t=\ell_{tl}+\ell_{ce}$
  \ELSE 
  \STATE
  {
  $\ell_t=\ell_{ce}$
  }
\ENDIF

\STATE \textbf{Step 2}: Generate pseudo-labels $\widetilde{y}_i$ for unlabeled images with test-time augmentations.

\STATE \textbf{Step 3}: Combine the ground truth labeled data with the pseudo labeled data with a proper ratio.

\STATE \textbf{Step 4}: Train a student network $\theta_{s}$ on the combined dataset with the video segmentation loss, the temporal loss and the distillation loss,
  $\ell_s=\lambda(\ell_{dis}+\ell_{tl})+\ell_{ce}$.
\end{algorithmic}
\end{algorithm}

\subsection{Temporal Consistency Knowledge Distillation}
\label{sec:tckd}

\noindent\textbf{Pair-wise-Frames Dependency.}  As shown in Figure~\ref{fig:overall} (c), following~\cite{liu2019structured}, we denote an attention (AT) operator to calculate the pair-wise similarity map $\mathbf{A}_{\mathbf{X}_1,\mathbf{X}_2}$ of two input tensors $\mathbf{X}_1,\mathbf{X}_2$, where $\mathbf{A}_{\mathbf{X}_1,\mathbf{X}_2} \in \mathbb{R}^{N \times N  \times 1}$ and $\mathbf{X}_1,\mathbf{X}_2 \in \mathbb{R}^{N  \times C} $. For the pixel $a_{ij}$ in $\mathbf{A}$, we calculate the cosine similarity between $\mathbf{x}_{1}^{i}$ and $\mathbf{x}_{2}^{j}$ from $\mathbf{X}_1$ and $\mathbf{X}_2$, respectively: 
\[ a_{ij} = {\mathbf{x}_1^i{}^\top \mathbf{x}_{2}^j}/{(\|\mathbf{x}_1^i\|_2\|\mathbf{x}_{2}^j\|_2)}.
\] 
This is an efficient and easy way to encode the correlations between two input tensors.

We encode the pair-wise dependency between the predictions of every two neighbouring frame pairs by using the AT operator, and %
obtain 
the similarity map $\mathbf{A}_{\mathbf{Q}_t,\mathbf{Q}_{t+k}}$, where $\mathbf{Q}_t$ is the segmentation map of frame $t$ and $a_{ij}$ of $\mathbf{A}_{\mathbf{Q}_t,\mathbf{Q}_{t+k}}$ denotes the similarity between the class probabilities on the location $i$ of the frame $t$ and the location $j$ of the frame $t+k$. If a pixel on the location $i$ of frame $t$ moves to location $j$ of frame $t+k$, the similarity $a_{ij}$ may be higher. Therefore, the pair-wise dependency can reflect the motion correlation between two frames.

We align the pair-wise-frame (PF) dependency between the teacher net $\mathsf{T}$ and the student net $\mathsf{S}$,
\begin{equation}
    \ell_{PF}(\mathbf{Q_t},\mathbf{Q_{t+k}})=\frac{1}{N^2}\sum_{i=1}^{N}\sum_{j=1}^{N}(a_{ij}^\mathsf{S}-a_{ij}^\mathsf{T})^{2},
\end{equation}
\noindent where $\forall a_{ij}^\mathsf{S} \in \mathbf{A}_{\mathbf{Q}_t,\mathbf{Q}_{t+k}}^\mathsf{S}$ and $
    \forall a_{ij}^\mathsf{T} \in \mathbf{A}_{\mathbf{Q}_t,\mathbf{Q}_{t+k}}^\mathsf{T}$.

\noindent\textbf{Multi-Frame Dependency.} As shown in Figure~\ref{fig:overall} (d), for a video sequence $\mathcal{I}=\{\dots \mathbf{I}_{t-1},\mathbf{I}_{t},\mathbf{I}_{t+1}\dots\}$, the corresponding feature maps 
$\mathcal{F}=\{\dots \mathbf{F}_{t-1},\mathbf{F}_{t},\mathbf{F}_{t+1}\dots\}$ are extracted from the output of the last convolutional block before the classification layer. Then, the self-s\-i\-m\-i\-l\-ar\-i\-t\-y map, $\mathbf{A}_{\mathbf{F}_t,\mathbf{F}_{t}}$, for each frame are calculated by using AT operator in order to: 1) capture the structure information among pixels, and 2) align the different feature channels between the teacher net and student net.

We employ a ConvLSTM unit to encode the sequence of self-similarity maps into an embedding $\mathbf{E}\in \mathbb{R}^{1 \times  D_e}$, where $D_e$ is the length of the embedding space.
For each time step, the ConvLSTM unit takes $\mathbf{A}_{\mathbf{F}_t,\mathbf{F}_{t}}$ and the hidden state which contains the information of previous $t-1$ frames as input; %
and outputs an embedding $\mathbf{E}_t$ along with the hidden state of the current time step. 

We align the final output embedding\footnote{The details of calculations in ConvLSTM is referred in~\cite{ConvLSTM}, and we also include the key equations in Section A.2 in Appendix.} at the last time step,   $\mathbf{E}^{\mathsf{T}}$ and  $\mathbf{E}^{\mathsf{S}}$ from $\mathsf{T}$ and $\mathsf{S}$, respectively. The output embedding encodes the relations of the whole input sequence, named multi-frame dependency (MF). The distillation loss based on multi-frame dependency is termed as: $    \ell_{MF}(\mathcal{F})=\left \|\mathbf{E}^{\mathsf{T}}-\mathbf{E}^{\mathsf{S}}\right \|_{2}^{2}$.

The parameters in the ConvLSTM unit are optimized together with the student net. To extract the multi-frame dependency, both the teacher net and the student net share the weight of the ConvLSTM unit. Note that there exists a model collapse point when the weights and bias in the ConvLSTM are all equal to zero. We clip the weights of ConvLSTM between a certain range and enlarges the $\mathbf{E}^{\mathsf{T}}$ as a regularization to prevent the model collapse.

 Finally, the distillation loss $\ell_{dis}$ in Equation~\ref{eq:all} can be written as $\ell_{dis}=\ell_{PF}+\ell_{MF}$.
\subsection{Motion Guided Temporal Consistency}
\label{sec:TL}

Training semantic segmentation networks independently on each frame of a video sequence often leads to undesired inconsistency. Conventional methods %
include 
previous predictions as an extra input, which introduces extra computational cost during inference.  
We employ previous predictions as supervised signals to assign consistent labels to each corresponding pixel
along the time axis. 
\begin{figure}[htbp]
    \centering 
    \includegraphics[width=0.4\textwidth]{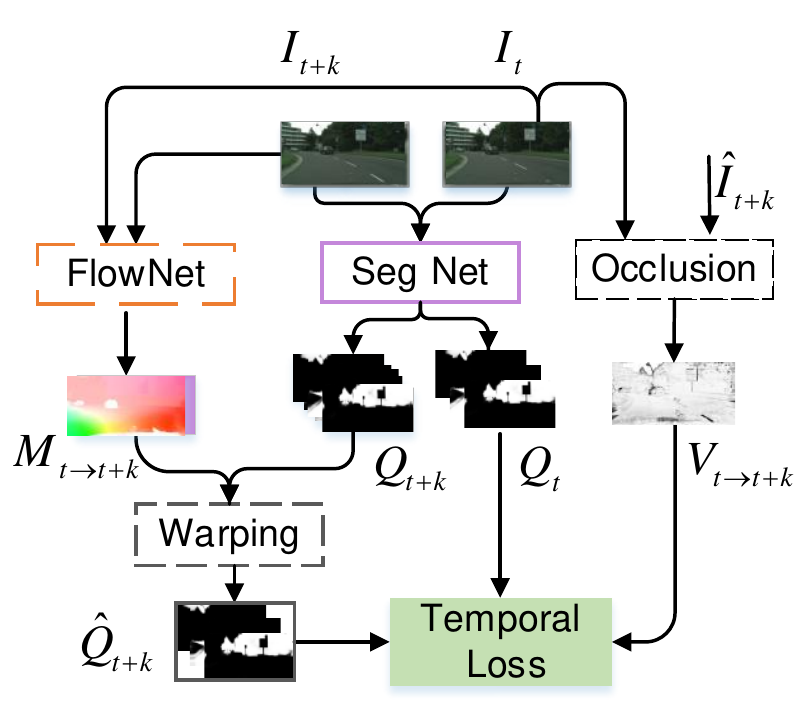}
\     \caption{\textbf{Temporal loss (TL)} encodes  the temporal consistency through motion constraints. Both the teacher net and the student net are enhanced by the temporal loss.}
    \label{fig:tl}
\end{figure}

 As shown in Figure~\ref{fig:tl}, for two input frames $\mathbf{I}_{t}$, $\mathbf{I}_{t+k}$ from time $t$ and $t+k$, we have:
\begin{equation}
    \ell_{tl}(\mathbf{I}_{t},\mathbf{I}_{t+k})=\frac{1}{N}\sum_{i=1}^{N}V_{t\Rightarrow t+k}^{(i)}\operatorname{KL}(\mathbf{q}_t^{i} \| \hat{\mathbf{q}}_{t+k\Rightarrow t}^{i} )
    \label{eq:tl}
\end{equation}
where $\mathbf{q}_t^{i}$ represents the predicted class probability at the position $i$ of the segmentation map $\mathbf{Q_{t}}$, $\operatorname{KL}(\cdot)$ is the Kullback$\textrm{-}$Leibler divergence
between two probabilities, and $\hat{\mathbf{q}}_{t+k\Rightarrow t}^{i}$ is the warped class probability from frame $t+k$ to frame $t$, by using a motion estimation network(e.g., FlowNet) $f(\cdot)$. Such an $f(\cdot)$ can predict the amount of motion changes in the $x$ and $y$ directions for each pixel: $f(\mathbf{I}_{t+k},\mathbf{I}_{t})=\mathbf{M}_{t \rightarrow t+k}$,
where $\delta i=\mathbf{M}_{t \rightarrow t+k}(i)$, indicating the pixel on the position $i$ of the frame $t$ moves to the position $i+\delta i$ in the frame $t+k$. Therefore,
the segmentation maps between two input frames are aligned by the motion guidance. An occlusion mask $\mathbf{V}_{t\Rightarrow t+k}$ is designed to remove the noise caused by the warping error: $\mathbf{V}_{t\Rightarrow t+k}=\exp(-\left | \mathbf{I}_{t}-
\hat{\mathbf{I}}_{t+k} \right |)$, where $\hat{\mathbf{I}}_{t+k}$ is the warped input frame. We employ a pre-trained optical flow prediction network as the motion estimation net in implementation. The frame gap $k$ could be any reasonable value according to the dataset. Here, we employ $k=1$.
We directly consider the temporal consistency during the training process through the motion guided temporal loss by constraining a moving pixel along the time steps to have a consistent semantic label.
Similar constraints are proposed in image processing tasks~\cite{lai2018learning,yao2017occlusion}, but 
are 
rarely discussed in semantic segmentation. We find that this 
straightforward temporal  consistency loss can improve the temporal consistency of single-frame models significantly.

\subsection{Video Instance Segmentation}
As shown in Algorithm~\ref{alg:semi}, the proposed training method can improve the temporal consistency for per-frame models. The proposed training scheme does not limit to semantic video segmentation, and it is easy to generalize to
other per-frame models, such as instance segmentation networks. 
Given a single input image $\textbf{I}_t$ on frame $t$ with the shape $W \times H$, the per-frame instance segmentation network can produce a binary mask with the shape $W \times H \times 1$ with a class from  $C$ predefined classes for each instance. To make a instance segmentation network work on the video sequence, we use a simple matching scheme by the weighted sum of the classification score, mIoU, and the location similarity to associate instances in adjacent frames following~\cite{yang2019video}. Given two input frame $\textbf{I}_t$ and $\textbf{I}_{t+1}$ at time steps $t$ and  $t+1$, the predicted mask for instance $n$ will be $\textbf{P}_{n,t}$ and $\textbf{P}_{n,t+1}$. The temporal  loss $\ell_{tl}$ follows Eq~\ref{eq:tl}
\begin{equation}
    \ell_{tl}(\textbf{I}_t,\textbf{I}_{t+1})=\sum_{n=1}^{N}\textbf{V}_{n,t\Rightarrow t+1}\left \|\textbf{P}_{n,t}-\hat{\textbf{P}}_{n,t+1\Rightarrow t}  \right \|_{2}^{2}. 
\end{equation}
Here, 
$\hat{\textbf{P}}_{n,t+1\Rightarrow t}$ is the warped trimap from frame $t+1$ to frame $t$, and 
$\textbf{V}_{n,t\Rightarrow t+1}$ is the occlusion mask based on the warping errors of RGB pixels.  We employ a recent instance segmentation method to predict the mask for instances of interest. For more details, refer to \cite{wang2020solov2}.

\section{Experiments: Semantic Video Segmentation}
\subsection{Implementation details}
\noindent\textbf{Dataset.}
Cityscapes~\cite{Cordts2016Cityscapes}
is collected for urban scene understanding
and contains $30$-frame snippets of the street scene with $17$ frames per second. The dataset contains $5,000$ high quality pixel-level finely annotated images at $20^{th}$ frame in each snippets, which are divided into $2,975$, $500$,  $1,525$ images for training, validation and testing. 
The dataset also contains 20K coarsely annotated video frames, namely `train-extra'. We generate pseudo labels on these video frames. The CamVid dataset~\cite{brostow2008segmentation} is an automotive dataset. It contains five different videos, which have ground truth labels every $30$ frame. Three train videos contain $367$ frames, while two test videos contain $233$ frames.

\noindent\textbf{Network structures.}
Different from keyframe based methods, which take several frames as input during %
inference,  
we apply our training methods to a compact segmentation model with per-frame inference. There are three main parts while training the system: a teacher net of PSPNet \cite{zhao2017pyramid} with a ResNet$101$ \cite{He2016DeepRL} trained with the temporal loss and the cross-entropy loss; A motion estimation network using a pre-trained FlowNetV2~\cite{flownet2-pytorch}; and a light-weight student network, including PSPNet18~\cite{liu2020efficient}, MobileNetV2~\cite{liu2020efficient}, HRNet-w18~\cite{liu2020efficient}, BiseNetV2~\cite{yu2020bisenet}, ESPNetV2~\cite{mehta2019espnetv2} and harDNnet~\cite{chao2019hardnet}.

\noindent\textbf{Random
sampling 
policy.}
To reduce the computational cost while training on video data, and make use of more unlabeled frames, we randomly sample frames %
before 
the ground-truth/pseudo labelled frames, named `frame\_f' and behind %
the labelled frame, named `frame\_b' to form a training triplet (frame\_f, labelled frame, frame\_b), instead of only using the frames right next to the labelled ones. The random %
sampling 
policy can take both long-term and short-term correlations into consideration and achieve better performance. Training on a longer sequence may
 show better performance with more expensive computation.

\noindent\textbf{Training.}
On Cityscapes, the same segmentation networks as the previous version~\cite{liu2020efficient} are trained by mini-batch stochastic gradient descent (SGD) for $200$ epochs. We sample $8$ training triplets for each mini-batch. The learning rate is initialized as $0.01$ and is multiplied by $(1-\frac{iter}{ max-iter})^{0.9}$.  We randomly cut the images into $769 \times 769$ as the training input. Normal data augmentation methods are applied during training, such as random scaling (from $0.5$ to $2.1$) and random flipping. Other off-the-shelf networks are trained following the original papers. On Camvid, we use a crop size of $640\times 640$. We use the official implementation of PSPNet in Pytorch\cite{semseg2019} and train all the network with $4$ cards of Tesla Volta $100$. Note that the the temporal consistency loss is only applied to the frame $t$ and  frame $t+1$. 

\noindent\textbf{Evaluation metrics.}
We evaluate our method on three aspects: accuracy, temporal consistency, and efficiency. The accuracy is evaluated by %
widely-used
mean Intersection over Union (mIoU) and pixel accuracy for semantic segmentation~\cite{liu2019structured}. We report the model parameters (\#Param) and frames per second (fps) to show the efficiency of employed networks. The inference speed is evaluated on a single GTX 1080Ti.  We follow \cite{lai2018learning} to measure the temporal stability of a video based on the mean flow warping error between every two neighbouring frames.  

We follow \cite{lai2018learning} to measure the temporal stability of a video based on the flow warping error between two frames. 

Different from \cite{lai2018learning}, we use the mIoU score instead of the mean square error to evaluate the semantic segmentation results,
\begin{equation}
    E_{warp}({\mathbf{Q}_{t-1},\mathbf{Q}_{t}})=\frac{\mathbf{Q}_{t}\cap \hat{\mathbf{Q}}_{t-1}}{\mathbf{Q}_{t}\cup \hat{\mathbf{Q}}_{t-1}},
\end{equation}
where $\mathbf{Q}_{t}$ represents for the predict segmentation map of frame $t$ and $\hat{\mathbf{Q}}_{t-1}$ represents for the warped segmentation map from frame $t-1$ to frame $t$.
We calculate a statistical average warp IoU on each sequence, and using an average mean on the validation set to evaluate the temporal stability:
\begin{equation}
    E_{warp}=\frac{1}{N}\sum_{i=1}^{N}\frac{\mathcal{Q}^{i}\cap \hat{\mathcal{Q}}^{i}}{\mathcal{Q}^{i}\cup \hat{\mathcal{Q}}^{i}},
\end{equation}
where $\mathcal{Q}=\{\mathbf{Q}_2,\dots,\mathbf{Q}_{T}\}$ and $\hat{\mathcal{Q}}=\{\hat{\mathbf{Q}}_1,\dots,\hat{\mathbf{Q}}_{T-1}\}$. $T$ is the total frames of the sequence and $N$ is the number of the sequence.
On Cityscapes~\cite{Cordts2016Cityscapes}, we random sample $100$ video sequence from the validation set, which contains $3000$ images to evaluate the temporal smoothness. { The sampled list has been public through our released code page. The temporal consistency of all compared methods in this work are evaluated on the same sampled subset.} On Camvid~\cite{brostow2008segmentation}, we evaluate the temporal smoothness of the video sequence `seq05' from the test set.

\begin{table*}[t]
\setlength{\abovecaptionskip}{10pt}
\caption{Accuracy and temporal consistency of the student net on Cityscapes validation set. SF: single-frame distillation methods, PF: our proposed pair-wise-frame dependency distillation method. MF: our proposed multi-frame dependency distillation method, TL: the temporal loss. PL: pseudo labels on extra video frames. The proposed distillation methods and temporal loss can improve both the temporal consistency and accuracy and they are complementary to each other.} 
\centering
		\setlength{\tabcolsep}{10pt}

\begin{tabular}{c|ccccc|c|c|c}

\hline
Scheme index&      SF&   PF      & MF        & TL & PL       & mIoU  & Pixel accuracy & Temporal consistency\\ \hline
  \multicolumn{9}{c}{Increasing Temporal Consistency}      \\ \hline
${a}$&&                    &           &                     && 69.79 & 77.18     & 68.50                \\ \hline

${b}$& &          \checkmark &           &                    & & 70.32 & 77.96     & 70.10                \\ \hline

${c}$& &                & \checkmark &                      & &    70.38    &        77.99    &    69.78                  \\ \hline

${d}$& & \checkmark & \checkmark          &                  &    & 71.16 &78.69    & 70.21                \\
\hline
     
${e}$&&                     &           & \checkmark &  &       70.67 & 78.46     & 70.46                \\ \hline

${f}$& & \checkmark &\checkmark           & \checkmark        &  &71.57           & 78.94 &70.61             
                    
\\\hline 
             \multicolumn{9}{c}{Combining with SF}      \\
\hline

${g}$&  \checkmark &         &&           &                     & 70.85 & 78.41     & 69.20                \\ \hline

${h}$ & \checkmark &           &                     & \checkmark&&71.36  &   78.64 &70.13                 \\ \hline

${i}$ & \checkmark & \checkmark & \checkmark          &\checkmark&  & 73.06 &80.75     & 70.56                \\
\hline 
  \multicolumn{9}{c}{Combining with PL}      \\
\hline
${j}$&             &           &           &           & \checkmark & 74.31 & 81.58          & 69.76    \\\hline
 ${k}$&             &           &           & \checkmark & \checkmark & 74.82 & 82.38          & 70.25                \\\hline
  ${l}$&           & \checkmark & \checkmark & \checkmark & \checkmark &\textbf{ 75.90} & \textbf{83.48 }         &\textbf{ 70.67 }       \\     

 \hline
\end{tabular}

\label{tab:abl}
\end{table*}

\subsubsection{Ablations}
Ablation experiments are conducted on Cityscapes with the PSPNet$18$ model. 

\noindent\textbf{Effectiveness of proposed methods.}
In this section, we verify the effectiveness of the proposed training scheme. Both the accuracy and temporal consistency are shown in Table~\ref{tab:abl}. We build the baseline scheme $a$, which is trained on every single labelled frame.  Then, we apply the proposed temporal consistency knowledge distillation terms: the pair-wise-frame dependency (PF) and multi-frame dependency (MF), separately, to obtain  the scheme ${b}$ and ${c}$. The temporal loss is employed in the scheme ${e}$. A  preliminary version of this  work~\cite{liu2020efficient}  employs the online single frame (SF) distillation on labeled images to further improve the segmentation accuracy, which requires extra memory during training.  Here, we propose to generate pseudo labels (PL) with test-time augmentation on unlabeled video sequences, which can be seen as an offline single-frame distillation. The detailed ablation results are shown in the scheme ${i}$, ${g}$ and ${k}$.

Compared with the baseline scheme, all the schemes can improve accuracy as well as temporal consistency.   
We also find when combine SF and PL together, the mIoU can be slightly improved, but the temporal consistency will be decreased. Thus, the SF is removed in this work.

To compare scheme ${g}$ with ${c}$ and ${d}$, one can see that the newly designed distillation scheme across frames can improve the temporal consistency to a greater extent. From the scheme ${e}$, it is observed that the temporal loss is most effective for the improvement of temporal consistency. 

To compare scheme ${g}$ and ${j}$, we can see that training on pseudo labels with video sequences can help the model learn more consistent results. To compare scheme ${f}$ with ${i}$, we can see that online single frame distillation methods~\cite{liu2019structured} can improve the segmentation accuracy but may harm the temporal consistency. In contrast, in the scheme ${l}$, off-line distillation on consecutive frames can help further improve the temporal consistency.

Compared with the preliminary conference version, the segmentation accuracy is improved from $73.06\%$ to $75.90\%$ in terms of mIoU while the temporal consistency is improved from $70.56\%$ to $70.67\%$.  We do not %
introduce 
any parameters or extra computational cost with per-frame inference. Both the distillation terms and the temporal loss can be seen as regularization terms, which can help the training process. Such regularization terms introduce extra knowledge from the pre-trained teacher net and the motion estimation network. Besides, performance improvement also benefits from temporal information and unlabelled data from the video.

\noindent\textbf{%
Impact of the random %
sampling 
policy.} We apply the random sampling (RS) policy when training with video sequence in order to make use of more unlabelled images, and capture the long-term dependency. Experiment results are shown in Table~\ref{tab:RS}. When the training images are not sufficient (without pseudo labels), the improvement of the temporal consistency is larger. By employing the random sampling policy, both the temporal loss and distillation terms can benefit from more sufficient training data in the video sequences in the next section. When employing more training data from the consecutive video frames, the random sampling policy can still have further improvement. We employ such a random sampling policy considering the memory cost during training. 

\begin{table}[htbp]
\setlength{\abovecaptionskip}{10pt}
\caption{Impact of the random sample policy. RS: random sample policy, TC: temporal consistency, TL: temporal loss, Dis: temporal consistency knowledge distillation 
terms, PL: pseudo labels. The proposed random sample policy can improve the accuracy and temporal consistency.}
\centering
	\setlength{\tabcolsep}{8pt}
\begin{tabular}{c|ccc}
\hline
Method         & RS        & mIoU  & TC   \\ \hline
PSPNet18 + TL  &           & 70.04 & 70.21      \\
PSPNet18 + TL  & \checkmark & 70.67 & 70.46 \\ \hline
PSPNet18 + Dis  &           & 70.94 & 69.98      \\ 
PSPNet18 + Dis  & \checkmark & 71.16 & 70.21\\ 
\hline
PSPNet18 + Dis+ TL+PL &           & 75.71 & 70.63 \\
PSPNet18 + Dis+ TL+PL & \checkmark & 75.90 & 70.67 \\ \hline
\end{tabular}
\label{tab:RS}
\end{table}

\begin{table}[b]
\setlength{\abovecaptionskip}{10pt}
	\setlength{\tabcolsep}{6.7pt}
\centering
\caption{Influence of the teacher net. TL: temporal loss. TC: temporal consistency. We use the pair-wise-frame distillation to show our design can transfer the temporal consistency from the teacher net.}
\begin{tabular}{l|ccc}
\hline
Method               & Teacher Model        & mIoU  & TC \\ \hline
PSPNet101            &   None                   & 78.84 & 69.71               \\ 
PSPNet101 + TL&          None            & 79.53 & 71.68               \\ \hline
PSPNet18             &       None               & 69.79 & 68.50                \\ 
PSPNet18             & PSPNet101            & 70.26 & 69.27                \\ 
PSPNet18             & PSPNet101 + TL & \textbf{70.32} & \textbf{70.10}                \\ 
\hline
\end{tabular}
\label{tab:teacher}
\end{table}

\noindent\textbf{Impact of the teacher net.}
The temporal loss can improve the temporal consistency of both cumbersome models and compact models. We compare the performance of the student net training with different teacher nets (\textit{i.e.}, with and without the proposed temporal loss) to verify that the temporal consistency can be transferred with our designed distillation term. The results are shown in Table~\ref{tab:teacher}. The temporal consistency of the teacher net (PSPNet$101$) can be enhanced by training with temporal loss by $1.97\%$. Meanwhile, the mIoU can also be improved by $0.69\%$.  By using the enhanced teacher net in the distillation framework, the segmentation accuracy is comparable ($70.26$ \textbf{vs.} $70.32$), but the temporal consistency has a significant improvement ($69.27$ \textbf{vs.} $70.10$), indicating that the proposed distillation methods can transfer the temporal consistency from the teacher net.

\subsubsection{Discussions}
\noindent \textbf{Correlations with model design works.} We focus on improving the accuracy and temporal consistency for real-time models by making use of temporal correlations. Thus, we do not introduce extra parameters during inference. We verify that our methods can generalize to different efficient backbones, \textit{e.g.}, ResNet18, MobileNetV2, and HRNet with ablation results.

A series of work~\cite{yu2020bisenet,chao2019hardnet,mehta2019espnetv2} focus on designing network structures for fast segmentation on single images and achieve promising results. 
We further conduct experiments on these structures and make them more suitable to process video sequences. { Besides, large models~\cite{zhao2017pyramid,zhu2019improving, chen2020naive} can achieve high segmentation accuracy with large backbones. For example, Zhu \emph{et al.}~\cite{zhu2019improving} achieve $81.4\%$ in terms of mIoU with ResNeXt50 as backbone. It has  low inference speed. Besides, the temporal loss is also effective when applying to large models, e.g., the temporal consistency and the accuracy of the PSPNet101 can be improved with temporal loss.}

\begin{table}[htb]
\caption{{Comparison on different metrics of temporal smoothness. Tsb is short for temporal stability~\cite{perazzi2016benchmark}.}}
\centering
\begin{tabular}{l|c|cc}
\hline
{Method}                & {mIoU}  & {Tsb}    & {TC}    \\
\hline
{PSPNet18}        & {69.79} & {0.0925} & {68.50} \\
{PSPNet18\_video} & {75.90} & {0.0892} & {70.67} \\
{PSPNet101}       & {78.84} & {0.0926} & {69.71} \\
{PSPNet101+TL}    & {79.53} & {0.0836} & {71.68}\\
\hline
\end{tabular}
\label{tsb}
\end{table}

\noindent \textbf{Smoothness and accuracy.} { Smoothness and accuracy are not independent of each other. For example, if we can predict every pixel correctly, the smoothness should achieve the best result. However, a stronger model does not naturally have higher temporal consistency. For example, the PSPNet101 achieves 78.8\% in terms of mIoU, which is higher than any other enhanced student nets, but the temporal consistency is only $69.7\%$ in terms of TC, which is very low. 

There are some previous metrics to evaluate the temporal smoothness, such as the temporal stability as used in~\cite{perazzi2016benchmark}, which evaluates temporal segmentation quality with a focus on boundaries. This metric is designed for video object segmentation, which only contains several objects in each frame. Converting the boundary of the mask into a collection of coordinate points has low efficiency on scene parsing datasets, like Cityscapes. Besides, some classes of stuff (\textit{e.g.}, road, sky) are hard to transfer. It takes around 13 hours to evaluate the temporal stability while 3 hours to evaluate the TC on Cityscapes. Thus, the proposed Temporal Consistency (TC) is suitable to evaluate the scene parsing dataset, like Cityscapes and CamVid. We report the comparison results between different evaluation metrics of temporal smoothness in Table~\ref{tsb}. Our proposed training scheme can improve the temporal smoothness in terms of both metrics.}

\begin{table*}[ht]
\setlength{\abovecaptionskip}{10pt}
\caption{{Comparison with recent efficient image/video semantic segmentation networks on three aspects: accuracy (mIoU,\%), smoothness (TC, \%) and inference speed (fps, Hz). Results are reported on the validation set of Cityscapes.  We sort all the methods based on three evaluation metrics. `R' is the ranking of the sum of three rankings (r1, r2, r3).}}
\centering
	\setlength{\tabcolsep}{5.2pt}
\begin{tabular}{l|l|c|cc|cc|cc|c}
\hline
\multirow{2}{*}{Method} & \multirow{2}{*}{Backbone} & \multirow{2}{*}{\#Params} & \multicolumn{6}{c|}{Cityscapes}& \multirow{2}{*}{R} \\\cline{4-9}
\multicolumn{1}{c|}{}                        &                           &                         & mIoU   &{ r1}  & TC&{r2}       & fps&{ r3}    \\\hline
             \multicolumn{9}{c}{Video-based methods: Train and infer on multi frames}      \\\hline

  CC~\cite{shelhamer2016clockwork}                                          & VGG16                   & -                       & 67.7&{24}     & 71.2&{3}     & 16.5&{15}&{15}      \\
                   DFF~\cite{zhu2017deep}                                         & ResNet101                   & -                       & 68.7&{23}     & 71.4   &{2} & 9.7&{17}     &{15}    \\
                                     GRFP~\cite{nilsson2018semantic}                                        &ResNet101                   & -                       & 69.4     & -&-       &- & 3.2&-         & -   \\
                   DVSN~\cite{xu2018dynamic}                                         & ResNet101                     & -                       & 70.3     & -& -& -       & 19.8     & -       & -  \\
                  { DVRL}~\cite{wang2020dynamic}                                         & {ResNet101}                     & -                       &{72.9}     & -& -& -       & -     & -       & -  \\

                                     Accel~\cite{jain2019accel}                                       & ResNet101/18                & -                       & 72.1     & {16}&70.3 &{9}    & 3.6 &{23}&{21}\\\hline
                                              \multicolumn{9}{c}{Single frame methods: Train and infer on each frame independently} \\
                                     \hline
  { NaiveS}~\cite{chen2020naive}                                         &  {Xception71}                    & -                       &{82.0}     & -& -& -       & -     & -       & -  \\                                     
  PSPNet~\cite{zhao2017pyramid} & ResNet101               & 68.1                    & 78.8&{2}     &69.7  &{14}   &  1.7&{24}    & {14}\\
  SKD-MV2~\cite{liu2019structured}                                     & MobileNetV2               & 8.3                     & 74.5  &{11}   &68.2    &{20} & 14.4 &{16}    & {19}       \\
                   SKD-R18~\cite{liu2019structured}                                   & ResNet18                    & 15.2                    & 72.7  &{15}   & 67.6&{21}     &8.0&{22}&{24}    \\
                   PSPNet18~\cite{zhao2017pyramid}                                    &  ResNet18                    & 13.2                    & 69.8  &{22}   & 68.5&{18}     & 9.5&{18}   &{ 24}       \\
                   HRNet-w18~\cite{SunXLW19,SunZJCXLMWLW19}                                   & HRNet                     & 3.9                     & 75.6  &{9}   & 69.1&{16}     & 18.9&{11}     & {11}       \\
                   MobileNetV2~\cite{Sandler2018MobileNetV2IR}                                 & MobileNetV2        & 3.2                     & 70.2 &{21}    & 68.4&{19}     & 20.8&{7}     &{19}    \\                                 
                             ESPNetV2~\cite{mehta2019espnetv2} & -& 2.70       & 66.0 &{25}   &   62.5&{24} & 83&{3}&{23} \\
                           harDNet~\cite{chao2019hardnet}&harDNet& 35.4         & 76.0  &{7}&  53.7&{25}  & 35&{5}&{13}  \\
                    BiseNetV2~\cite{yu2020bisenet}&-& 21.2      & 71.3&{18}  &  67.2&{23}  & 156&{1}&{15}\\
                   \hline
                                           
                           \multicolumn{9}{c}{Train on multi frames and infer on each frame independently} \\ \hline 

Teacher Net & ResNet101               & 68.1                    & 79.5 &{1}    &71.7&{1}     &  1.7&{24}&{4}\\
  PSPNet18+TL                                  & ResNet18              & 13.2                    & 71.1   &{19}  & 70.0 &{12}    & 9.5 &{18}&{22}       \\
                   PSPNet18~\cite{liu2020efficient}                                    &  ResNet18                    & 13.2                    & 73.1  &{13}   & 70.6&{5}     & 9.5&{18}&{11}       \\
                   PSPNet18\_Video                                    &  ResNet18                    & 13.2                    & 75.9  &{8}   & 70.7&{4}     & 9.5&{18}&{6}       \\
                   HRNet-w18+TL                                   & HRNet                     & 3.9                     & 76.4   &{6}  & 69.6 &{15}    & 18.9     & {11}&{7}       \\
                   HRNet-w18~\cite{liu2020efficient}                                   & HRNet                     & 3.9                     & 76.6  &{5}   & 70.1&{11}     & 18.9&{11}&{5}       \\
                \textbf{HRNet-w18\_Video  }                                 & HRNet                     & \textbf{3.9}                     & \textbf{78.6} &{\textbf{4}}  & \textbf{70.6} &{\textbf{5}}    & \textbf{18.9}&{\textbf{11}}&{\textbf{2}}       \\
                   MobileNetV2+TL                                 & MobileNetV2           & 3.2                     & 70.7&{20}     & 70.4&{7}     & 20.8  &{7}   &{10}    \\
                   MobileNetV2~\cite{liu2020efficient}                                 & MobileNetV2      & 3.2                     & 73.9 &{12}    & 69.9&{13}     & 20.8    &{7}&{7}   \\
           \textbf{MobileNetV2\_Video }                                & MobileNetV2      & \textbf{3.2}                     & \textbf{75.5} &{\textbf{10}}    & \textbf{70.4}    &{\textbf{7}} & \textbf{20.8}&{\textbf{7}}&{\textbf{3}}   \\

{BiseNetV2\_Video}&-& 21.2 &  72.8  &{14}   & 68.7 &{17}  & 156&{1}&{9} \\
\textbf{{harDNet\_Video}}&harDNet& \textbf{35.4}   & \textbf{78.7} &{\textbf{3}}&  \textbf{70.3}&{\textbf{9}}  & \textbf{35}&{\textbf{5}}&{\textbf{1}}\\
{ESPNetV2\_Video}& -& 2.70 & 71.5&{17} &    67.3&{22}& 83&{3}&{15}   \\               
                 \hline 
\end{tabular}
\label{SOTA}
\end{table*}

\subsection{Results on Cityscapes} 

\noindent\textbf{Comparison with single-frame based methods.}
Single-frame methods are trained and inferred on each frame independently. Directly applying  such methods to video sequences %
often
produces inconsistent results. We first apply our training schemes to the same single-frame semantic segmentation networks as the previous version~\cite{liu2020efficient}, and achieve significant improvements. Metrics of 
mIoU, temporal consistency, inference speed, and model parameters are shown in Table~\ref{SOTA}.

As shown in Table~\ref{SOTA},
the proposed training scheme 
works well with a few
compact backbone networks (\textit{e.g.}, PSPNet18, HRNet-w18, and MobileNetV2). Both temporal consistency and segmentation accuracy can be improved 
using 
the temporal information among frames.

We also compare our training methods with the single-frame distillation method~\cite{liu2019structured}. According to our observation, GAN based distillation methods proposed in~\cite{liu2019structured}%
can
produce inconsistent results. For example, with the same backbone ResNet18, training with the GAN based distillation methods (SKD-R18) 
achieves higher mIoU: $72.7$ vs.\  $69.8$, and a lower temporal consistency: $67.6$ vs.\  $68.5$ compared with the baseline PSPNet18, which is trained with cross-entropy loss on every single frame. 
With the proposed training schemes using the temporal information among frames, noted as `PSPNet18\_Video', both the temporal consistency and accuracy are improved by a large margin.
Note that we also employ a smaller structure of the PSPNet with half channels than in~\cite{liu2019structured}.

As real-time semantic image segmentation methods focus on achieving higher accuracy with fewer computational costs, the model capacity is limited, and the results on video sequences are very inconsistent. As Table~\ref{SOTA} shows, when applying to off-the-shelf real-time semantic image segmentation models, the temporal consistency will improve significantly. Besides, the segmentation accuracy also boosts. We do not change the network structure of the off-the-shelf models. Therefore, with our improved model parameters, the networks can achieve the same inference speed as reported in original papers. The proposed training scheme has proven effective in multiple different network structures. Although the network structures of real-time per-frame models will achieve new state-of-the-art performance in the future, the correlations among frames will often help boost temporal consistency and accuracy.

\begin{figure}[t]
    \centering 
    \includegraphics[width=0.5\textwidth]{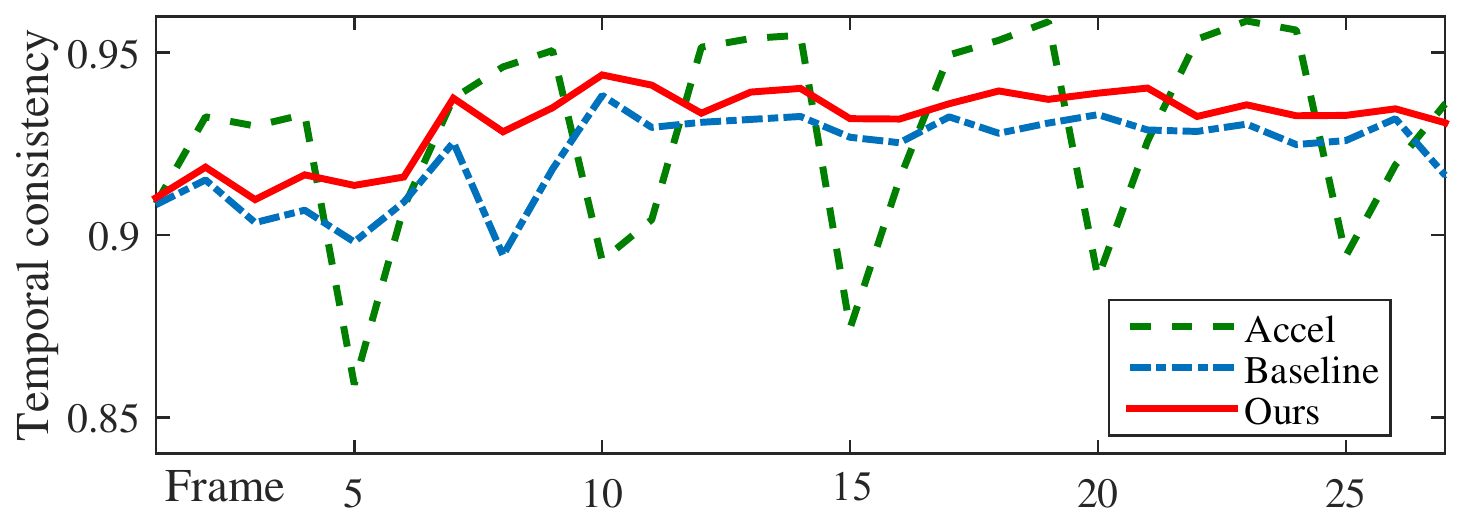}
    \caption{The temporal consistency between neighbouring frames in one sampled sequence on Cityscapes. The keyframe based method Accel 
    shows
    severe jitters between keyframes and others.}
    \label{fig:tc_one}

\end{figure}{}

\noindent\textbf{Comparison with video-based methods.} Video-based methods are trained and inferred on multi frames, we list current methods including keyframe based methods: CC~\cite{shelhamer2016clockwork}, DFF~\cite{zhu2017deep}, DVSN~\cite{xu2018dynamic}, Accel~\cite{jain2019accel}  and multi-frame input method: GRFP~\cite{nilsson2018semantic} in Table~\ref{SOTA}. The compact networks with per-frame inference can be more efficient than video-based methods. Besides, with per-frame inference, semantic segmentation networks have no unbalanced latency and can handle every frame independently. 
Table~\ref{SOTA} shows the proposed training schemes on simple network structures can achieve a better trade-off between the accuracy and the inference speed compared with other state-of-the-art semantic video segmentation methods, especially the MobileNetV2 with the fps of $20.8$ and mIoU of $75.5\%$.  With the strong harDNet, the mIoU can achieve $78.7\%$ with $35$ fps on a GTX $1080$Ti while the fastest video-based methods can only achieve 19.8 fps on the same mobile devices. Besides, the temporal consistency of harDNet can also achieve more than $70\%$, { and achieve the best ranking in terms of three rankings.} Although keyframe methods can achieve a high average temporal consistency score, the predictions beyond the keyframe are of low quality. Thus, the temporal consistency will be very low between keyframe and non-key frames, as shown in Figure~\ref{fig:tc_one}. The high average temporal consistency score is mainly from the low-quality predictions on non-key frames. In contrast, our method can produce stable segmentation results on each frame.

\begin{figure*}[tbp]
    \centering 
    \includegraphics[width=1.0\textwidth]{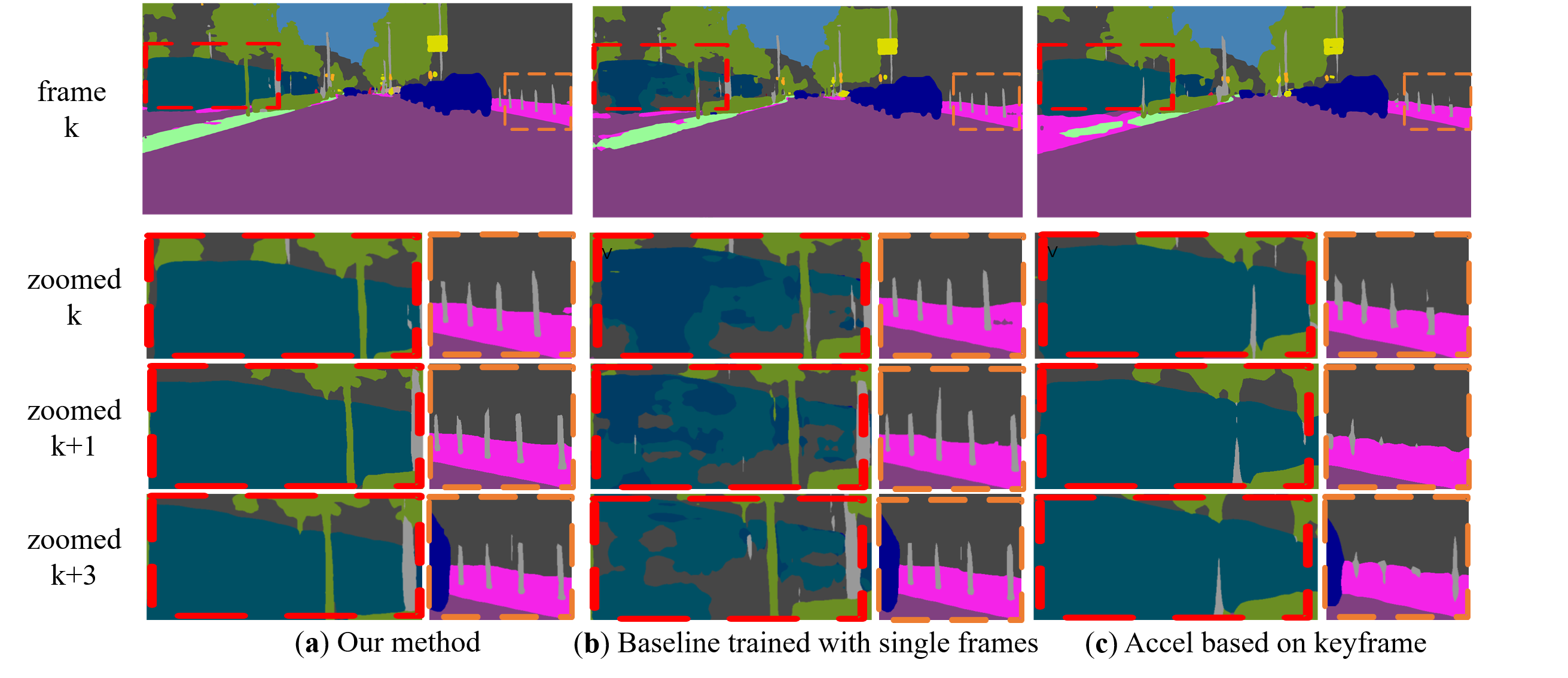}
    \caption{\textbf{Qualitative outputs.}  (\textbf{a}): PSPNet18, training on multi frames and inferring on each frame. (\textbf{b}): PSPNet18, training and inferring on each frame. (\textbf{c}): Accel-18~\cite{jain2019accel}, training and inferring on multiple frames. The keyframe is selected in every five frames. For better visualization, we zoom the region in the red and orange box. The proposed method can give more consistent labels to the moving train and the trees in the red box. In the orange boxes, we can see our methods have similar quantity results in each frame while the keyframe based methods may generate worse results in the frame (e.g., $k+3$) which is far from the keyframe (i.e., $k$).}
    \label{fig:city_vis}
\end{figure*}
\begin{table*}[htb]
\centering
\caption{Accuracy (mIoU, \%) and temporal consistency  (TC, \%) for each class on Cityscapes. Baseline: PSPNet18 trained on each frame independently. Ours: PSPNet18 trained with temporal loss and distillation items.}
	\setlength{\tabcolsep}{5pt}
\begin{tabular}{l|l|c|c|c|c|c|c|c|c|c|c}
\hline
\multicolumn{2}{c|}{Class Name}                                           & \multicolumn{1}{c|}{road}   & \multicolumn{1}{c|}{sidewalk} & building & wall   & fence  & pole   & tra. light & tra. sign & vegetation & terrain  \\\hline
\multicolumn{1}{c|}{\multirow{2}{*}{mIoU}} & \multicolumn{1}{c|}{Baseline} & \multicolumn{1}{c|}{97.0} & \multicolumn{1}{c|}{78.7}   & 90.1   & 41.8 & 54.7 & 50.3 & 63.6        & 72.0       & 90.8     & 60.0   \\
\multicolumn{1}{c|}{}                      & Ours                         & \textbf{97.2}                     & \textbf{79.4}                       & \textbf{91.0}   & \textbf{49.8} & \textbf{57.4} &\textbf{ 53.1} & \textbf{67.0}        & \textbf{73.6 }      & \textbf{91.0}     & 60.0  \\\hline
\multicolumn{1}{c|}{\multirow{2}{*}{TC}}                       & Baseline                     & 97.2                     & 80.2                       & 91.2   & \textbf{50.0} & 62.1 & 42.6 & 47.2        & 52.6       & 91.7     & 72.0  \\
                                          & Ours                         &\textbf{ 97.7}                     & \textbf{81.4}                       & \textbf{91.6}   & 49.6 & \textbf{62.6} & \textbf{43.9} & \textbf{48.5 }       & \textbf{53.2}       & \textbf{91.9}     & \textbf{73.3}  \\
                                          \hline
                                          
\multicolumn{2}{l|}{Class Name}                                           & sky                        & person                       & rider    & car    & truck  & bus   & train         & motorbike  & bicycle    & mean     \\\hline
\multicolumn{1}{c|}{\multirow{2}{*}{mIoU}}                                   & Baseline                     & 92.8                     & 75.8                       & 52.7   & 91.6 & 61.4 & 77.1 & 56.9        & 46.9       & 71.8     & 69.8  \\
                                          & Ours                         & \textbf{93.1}                     & \textbf{77.1}                       &\textbf{ 57.1}   & \textbf{92.1} & \textbf{65.5} &\textbf{ 82.2} & \textbf{73.1}        & \textbf{55.6}       & \textbf{72.8}     & \textbf{73.1}   \\\hline
\multicolumn{1}{c|}{\multirow{2}{*}{TC}}                      & Baseline                     & 92.8                     & 68.7                       & 28.7   & 86.4 & 74.8 & 78.5 & 55.5        & 55.9       & 73.7     & 68.5  \\
                                          & Ours                         & \textbf{93.0}                     &\textbf{ 69.6}                       & \textbf{30.1}   & \textbf{87.0} & \textbf{76.3} & \textbf{82.2} & \textbf{76.4 }      &\textbf{ 57.5}       & \textbf{74.9}     & \textbf{70.6}\\ 
                                          \hline
\end{tabular}
\label{fig:classes}
\end{table*}
\noindent\textbf{Qualitative visualization.}
Qualitative visualization results are shown in Figure~\ref{fig:city_vis}, in which, the keyframe-based method Accel-18 will produce unbalanced quality segmentation results between the keyframe (e.g., the orange box of $k$) and non-key frames (e.g., the orange box of $k+1$ and $k+3$ ), due to the different forward-networks it chooses. By contrast, ours can produce stable results on the video sequence because we use the same enhanced network on all frames. Compared with the baseline method trained on single frames, our proposed method can produce more smooth results, e.g., the region in red boxes. The improvement of temporal consistency is more clearly shown in the video comparison results.  Moreover, we show a case of the temporal consistency between neighbouring frames in a sampled frame sequence in Figure~\ref{fig:tc_one}. Temporal consistency between two frames is evaluated by the warping pixel accuracy. The higher, the better. The keyframe based method will produce jitters between keyframe and non-key frames, while our training methods can improve the temporal consistency for every frame. The temporal consistency between non-key frames are higher than our methods, but the segmentation performance is lower than ours.

\noindent\textbf{Results on each class.}
We compare our method with the baseline methods of PSPNet18 in terms of the accuracy and temporal consistency of each class on Cityscapes. The results are shown in Table~\ref{fig:classes}. For the moving objects with regular structures, e.g. `train', `bus', both segmentation accuracy and temporal consistency are improved significantly. For the `road', `sidewalk' and `terrain', the temporal consistency is also improved although the accuracy only has limited improvements.

\subsection{CamVid}
We provide additional experiments on CamVid.
We use %
MobileNetV2 as the backbone in the PSPNet.  In Table~\ref{SOTA}, the segmentation accuracy, and the temporal consistency are improved compared with the baseline method. We also outperform current state-of-the-art semantic video segmentation methods with a better trade-off between the accuracy and the inference speed. We use the pre-trained weight from cityscapes following VideoGCRF~\cite{chandra2018deep}, and achieve better segmentation results of $78.2$ vs. $75.2$. VideoGCRF~\cite{chandra2018deep} can achieve $22$ fps with $321\times321$ resolution on a GTX $1080$ card.
We can achieve $78$ fps with the same resolution.
The consistent improvements on both datasets verify the value of our training schemes for real-time semantic video segmentation.

\begin{table}[ht]
	\caption{{Accuracy and inference speed on \textbf{CamVid}. We use the MobileNetV2 as the backbone of the PSPNet. Compared methods include DFF~\cite{zhu2017deep}, Accel-18~\cite{jain2019accel}, GRFP~\cite{nilsson2018semantic}, VideoGCRF~\cite{chandra2018deep}, SKD-R18\cite{liu2019structured}, PSPNet\cite{zhao2017pyramid} }}
	\centering
		\footnotesize
	\label{tbl:cam}%
		\setlength{\tabcolsep}{5pt}
	\begin{tabular}[t]{lcccc}
\hline
		Model &  mIoU & TC&fps \\
		\hline
		\multicolumn{4}{c}{Video-based methods} \\
		\hline
		DFF & 66.0 &  78.0&16.1\\
		Accel-18 & 66.7 &76.2& 7.14 \\
			GRFP &66.1 &-& 6.39 \\
		VideoGCRF & 75.2 & - &-\\ \hline
	\multicolumn{4}{c}{Single frame methods} \\
			\hline
		SKD-R18&72.3&75.4&13.3\\
		MobileNetV2&74.4&76.8&27.8\\
		PSPNet&77.6&77.1&4.1\\
		\hline
		\multicolumn{3}{c}{Ours} \\
		\hline
		Teacher Net&79.4&78.6&4.1\\
        MobileNetV2+TL&76.3&77.6&\textbf{27.8}\\
        MobileNetV2\_video&78.2&77.9&\textbf{27.8}\\
\hline
	\end{tabular}
	\vspace{-1.5em}
\end{table}

\subsection{{300VW-Mask}}
{\textbf{Implementation details.} Both Cityscape and CamVid are outdoor datasets focus on the road scene. To test the generalization ability of our method, we conduct experiments with a new application, face mask segmentation, on the 300VW-Mask dataset~\cite{wang2019face}. The 300VW-Mask dataset is sampled from 300 videos in the wild, which contains 114 videos taken in unconstrained environments. The per-pixel annotations include one background class and four foreground classes: facial skin (FC), eyes, outer mouth (OMT), and inner mouth (IMT). 619/58/80 of 1-s sequences (30fps) selected from 93/9/12 videos are used for training/validation/testing, which contain 18570/1740/2400 face images in total. We train the model using the crop size of $385 \times 385$ for 80 epochs with $16$ images per batch. PSPNet18 is employed as our per-frame model. The teacher net with ResNet50 is trained with temporal loss. Following the previous work~\cite{wang2019face}, all the metrics are calculated on the original image resolution, and the mIoU is calculated without the background class.

\noindent\textbf{Experiment settings.} Different from the previous two datasets, 300VW-Mask has annotations on each frame in the 1-s sequence. Thus, we conduct experiments under two different settings. For the supervised setting, we train the per-frame model on the whole labeled dataset with and without learning from the frames. For the semi-supervised setting, we sampled $10\%$ of the labeled data, and train the per-frame models (both the teacher and student network) on the labeled data. Then, the pseudo labels are generated for other unlabeled frames. 

\noindent\textbf{Results.} Experiment results are shown in Table~\ref{300vw}. It is clearly shown that learning from the video sequence during training can significantly improve the accuracy and the smoothness of the per-frame model. We achieve $67.41\%$ in terms of mIoU with only $10\%$ of the labeled data, outperforming state-of-the-art methods with the same training splits. Our proposed method can have a larger improvement under the semi-supervised setting, demonstrating that part of the gain comes from making use of unlabeled data as analyzed before. The student network can achieve better performance as the pseudo labels are generated by test-time augmentation, which is stronger than the single-scale teacher net.
}

\begin{table*}[htp]
\centering
\caption{{Experiments results on 300VW-Mask~\cite{wang2019face}. The temporal stability (Tsb), temporal consistency (TC) and Insertion-over-Union (IoU) for each classes are reported. mIoU is calculated without the background class. The compared methods can be referred in~\cite{wang2019face}.}}
\begin{tabular}{l|c|c|c|c|c|c|c|c|c}
\hline
{Method}          & {mIoU}  & {FC}  & {Eyes}  & {OMT}  & {IMT}  & {BG}&{TC$\uparrow$}&{Tsb$\downarrow$}&{Backbone} \\ \hline
\multicolumn{10}{c}{{Train on the Whole Labeled Dataset}} \\ \hline
{FME} &{63.76}&{90.58}&{57.89}&{62.78}&{43.79}&{94.36}&-&-&{ResNet50}\\
{Face tracker}& {60.09}& {88.77}& {50.01}&{ 61.04}&{ 40.56}& {97.71}&-&-&-\\
{DeeplabV2}&{ 58.66}& {90.55}& {50.19}& {58.58}& {35.31}& {94.38}&-&-&{VGG16}\\
{FCN-VGG16}& {55.71}& {91.12}& {44.18}& {58.60}& {28.95}& {94.87}&-&-&{VGG16}\\\hline
{Teacher}           &{67.61}&    {91.04}   &{61.44}     &{\textbf{63.5}}       & {\textbf{54.45}}     &   {92.29}  &{71.31} &\textbf{{0.008013}}& {ResNet50}   \\ 
{Baseline}          &{59.04}   &{89.71}     & {50.82}      &{53.23}      &{42.38}      &    { 90.06}    &{69.72}&{0.008122}& {ResNet18} \\ 
{Baseline\_video}   &{\textbf{67.62}}       & {\textbf{91.33}}   &{\textbf{64.00}}       &{62.63}     &{52.53}      &\textbf{{92.56}}&{\textbf{71.33}}&{0.008025}&{ResNet18}    \\ \hline
\multicolumn{10}{c}{{Train on the 10\% Labeled Dataset}}  \\ \hline
{Teacher}         &    {65.74}   &  {91.20}   &{59.91}& {62.00}     & {49.88}&{92.51} &{70.77}&{0.00827}& {ResNet50}  \\ 
{Baseline}       &{50.81} &{88.97}       & {50.03}    &  { 59.59}    &    {46.51}  &  {90.26}    & { 69.09}&{0.01098}& {ResNet18} \\ 
{Baseline\_video}  &  {67.41}     &   {90.97}  &  {61.62}    &  {63.05}  & {53.98}    &  {92.17}&{71.27}&{0.00839}& {ResNet18} \\ \hline
\end{tabular}

\label{300vw}
\end{table*}

\begin{figure}
  \centering
 \includegraphics[width=0.47\textwidth]{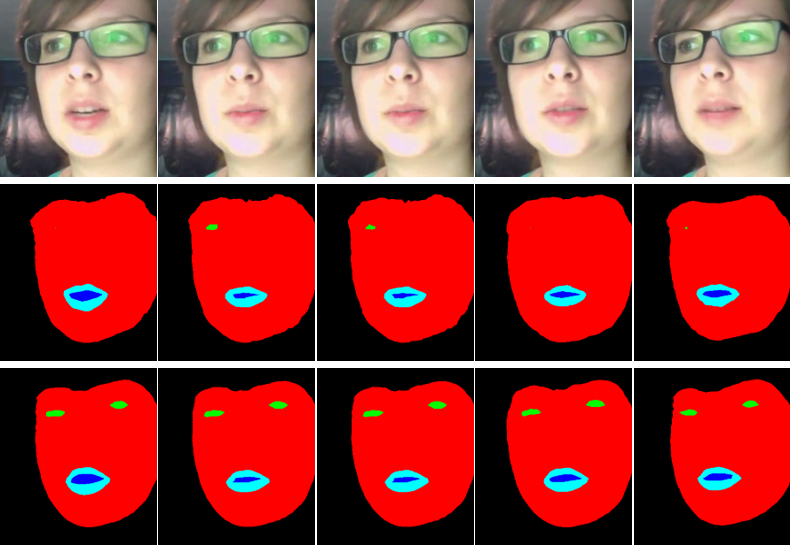}
  \caption{{Visualization results on 300VW-Mask.
  \textbf{First row}: 
  Input frames; 
  \textbf{Second row}: Segmentation results from the baseline under semi-supervised settings. 
  \textbf{Third row}:  Segmentation results from ours under semi-supervised settings.} 
  }
  \label{fig:300vw}
\end{figure}

\section{Experiments: Video Instance Segmentation}
{
In this section, we apply the proposed training scheme to the per-frame instance segmentation network. As a large teacher net is hard to obtain and the online knowledge distillation cost too much training memory in instance segmentation, we only employ the temporal loss and the offline distillation in this section.

First, we evaluate the effectiveness of our training scheme on the public dataset  YouTube-VIS~\cite{Yang_2019_ICCV}. We improve the accuracy of the per-frame baselines by considering the temporal information during training. As the proposed temporal loss and the pseudo labels do not depend on the human annotated labels, we propose a novel application of instance-level portraits matting in video sequences on the Internet by predicting instance-level trimaps.

Existing large scale matting datasets often consist of single images without consecutive frames and instance identities. We generate pseudo trimaps for each portraits in video sequences collected from the Internet. A teacher model is trained on MSCOCO for image instance segmentation. We dilate and erode the edge of the binary mask for each instance in consecutive frames to generate pseudo trimaps in collected videos. We also combine the pseudo labels with ground-truth trimaps from a Human Matting dataset  to get more supervision on the fine-grained edges. 
}

{ SOLOv2~\cite{wang2020solov2} is employed as the per-frame model in this section. It is a one-stage instance segmentation network, which contains a classification branch and a mask prediction branch. We initialize the network with the pre-trained weight on MSCOCO~\cite{lin2014microsoft}, containing $80$ classes of common objects. To train the model on public dataset YouTube-VIS~\cite{Yang_2019_ICCV}, we change the output channel of the classification branch from $80$ to $40$, and fine tune on the YouTube-VIS. To apply to the portrait matting in the wild, we change the output of the mask branch, by predicting a three-channel trimap in stead of the binary mask. Then, we fine tune the mask branch on the combined dataset.}

\subsection{Evaluation on YouTube-VIS}
\noindent\textbf{Implementation details.} { YouTube-VIS~\cite{Yang_2019_ICCV} contains 2238 training, 302 validation and 343 test video clips. Each video of the dataset is annotated with per pixel segmentation mask, category and
instance labels. The object category number is 40. The evaluation metrics are average precision (AP) and average recall (AR).  The initial learning rate is set to
0.05 and decays with a factor of 10 at epoch 8 and 11. The network is trained with 16 images per batch for 12 epochs. As YouTube-VIS already has accurate per-frame labels, we do not generate pseudo labels in this experiments. The temporal loss is employed to help the per-frame model learn from the video sequence. 

\noindent\textbf{Results.} We submit the testing results to the test server\footnote{\url{https://competitions.codalab.org/competitions/20128\#results}} of YouTube-VIS, and report the results in Table~\ref{tab:vis}. With the help of learning from the video sequence, the accuracy of our simple per-frame baselines is improved. SOLOv2\_video can achieve competitive results compared with previous methods.}

\begin{table}[htb]
\centering
\caption{{ Evaluation results on the test set of Youtube-VIS. Compared methods include MaskTrack-RCNN~\cite{Yang_2019_ICCV}, STEm-Seg~\cite{athar2020stem} and SOLOV2~\cite{wang2020solov2}.  }}
\setlength{\tabcolsep}{3pt}
\begin{tabular}{l|c|c|c|c|c}
\hline
          {Method}& {mAP}                  & {AP50}                 & {AP75}                 & {AR1}  & {AR10} \\ \hline
{MaskTrack-RCNN} & {32.3}                 & {53.6}                 & {34.2}                 & {33.6} & {37.3} \\
{STEm-Seg}&{34.6} &{55.8}& {37.9} &{34.4}& {41.6}\\\hline
{SOLOV2}         & {31.2}                 & {53.7}                 & {33.5}                 & {31.0} & {35.1} \\
{SOLOV2\_video}   & {\textbf{35.1}}                 & {\textbf{52.8}}                 & {\textbf{40.6}}                 & {\textbf{35.5}} & {\textbf{42.1}} \\
\hline
\end{tabular}
\label{tab:vis}
\end{table}

\subsection{Application on Trimap Prediction in the Wild}

\noindent\textbf{Dataset.} Video Dataset: we collect video sequences from YouTube with the keywords `dancing'  and `TikTok dancing'. The collected video sequences are from $200$ different scenes, and each frame containing single or multiple persons. We also collect $5$ video scenes with $10$ different instances ($1 \sim  3$ instances in each scene) as the test video sequence. We finally collect $10^4$ frames in all video sequences as the input images for generating pseudo labels.  Human Matting Dataset: The training set has $931$ foregrounds. There are $23275$ training images, $100$ validation images, and $400$ test images, including $931$/$40$/$80$ foregrounds. The ground-truth trimaps for training are generated from the ground truth alpha matte following~\cite{lu2019indices}. 

\noindent\textbf{Training.} The network is trained on the combined dataset of the human matting dataset and the pseudo labels of videos in the wild.  The training image size is $960\times 640$ with a batch size of $8$ images. We use the standard Stochastic Gradient Decent (SGD) for training. The trimap generation module is trained for $12$ epochs with an initial learning rate of $0.0005$, which is then divided by $10$ at $7$th epoch. For each frame with the pseudo label, we use consecutive pairs of frames during training to apply the temporal loss between these two consecutive frames. 

\subsubsection{Visualization results.}
\begin{figure*}
\centering  %
\includegraphics[width=0.9\textwidth]{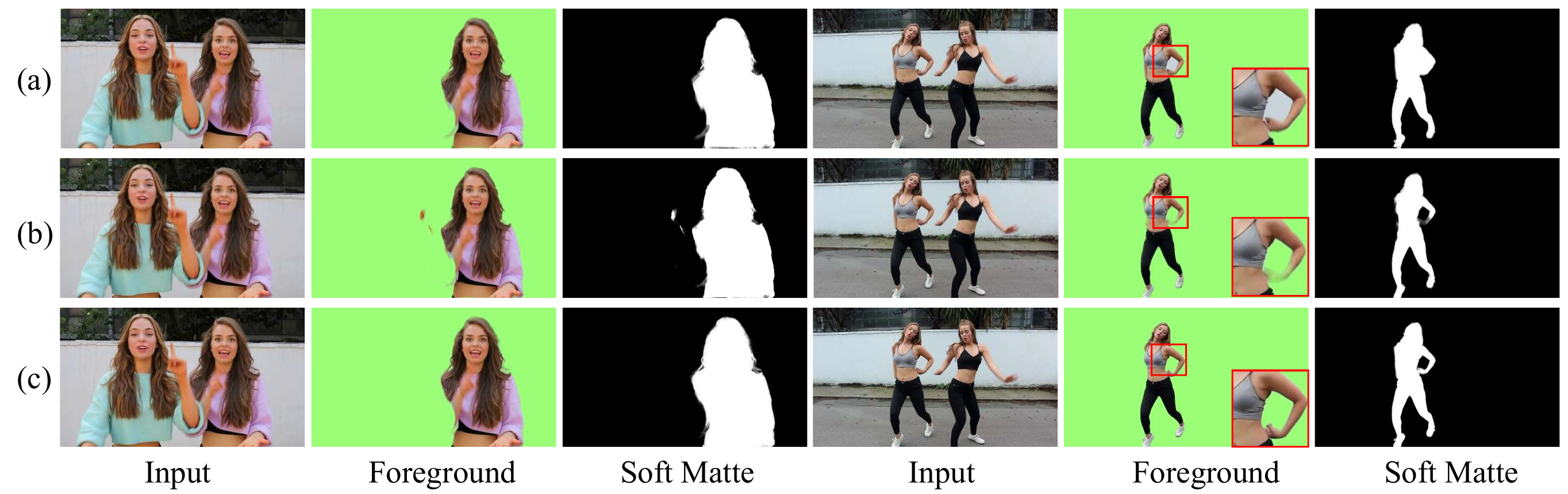}
\caption{\textbf{Foreground instances and soft mattes with different trimaps}: (a) SOLOv2~\cite{wang2020solov2} trained on the COCO dataset, and the trimap is generated by the dilated and eroded operation. This is the pseudo labels that we used. IndexNet is applied to the generated trimaps. The background between the arms and the body can not be distinguished by the model.
(b) Trimaps are predicted by the trimap prediction branch which is fine-tuned on the human matting dataset. The human matting dataset does not contain instance identity information, which may confuse the segmentation of adjacent objects. (c) Trimaps are predicted by the 
trimap prediction branch which is fine tuned on both the human matting dataset and the pseudo labels.
}
\label{Fig.dataset}
\vspace{-1em}
\end{figure*}

\noindent \textbf{Learning from pseudo labels.} { For better visualize the results, we apply a matting model, IndexNet~\cite{lu2019indices}, on the predicted trimap sequence.} We show one of the foreground instances and the soft mattes with different models on sampled frames of the { test} videos in Figure~\ref{Fig.dataset}. The combination of the pseudo labels with the ground-truth matting labels improves the trained model on the adjacent objects and the background area inside the human mask.

\begin{figure}[htp]
\centering  %
\includegraphics[width=0.45\textwidth]{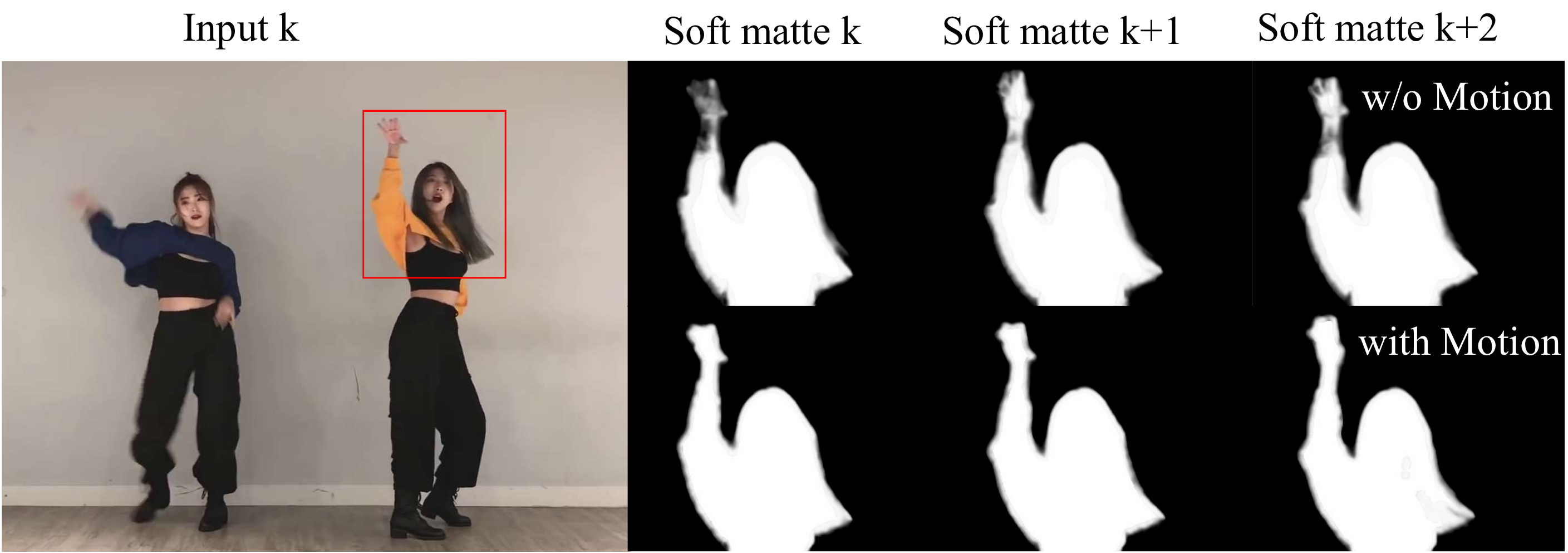}
\caption{\textbf{Role of the temporal loss.}
We show the results with three consecutive frames ($k$, $k+1$, $k+2$). With the motion relations between frames, the model can produce more consistent soft matting results.}
\label{Fig.mot}
\vspace{-1.9em}
\end{figure}

\noindent \textbf{Effective of the temporal loss.} 
Figure~\ref{Fig.mot} shows the ablation results with and without the temporal loss on the video sequence. The results with three consecutive frames ($k$, $k+1$, $k+2$) show that with the correlations among frames, the model can produce more consistent soft mattes. In the video sequence, there are some frames with clear foreground and some frames with motion blur. The blur frames benefit from the supervision from the clear frames. 

\begin{table}[htb]
\centering
\setlength\tabcolsep{4pt}
\caption{{ Zero-shot testing results on the person subset of Video Matting dataset~\cite{Erofeev2015}. TC: temporal consistency. Tsb: temporal stability.}}
\begin{tabular}{l|cccc|cc}
\hline
{Method}             & {BG}    & {Tri}   & {FG}    & {mIoU$\uparrow$}  & {TC $\uparrow$}& {Tsb$\downarrow$} \\
\hline
{SOLO}       & {0.79} &{ 0}     &{ 0.78} & {0.52} &  -  &    - \\\hline
{w/o TL, PL}  & {0.86} & {0.15} & {0.84} & {0.61} &  {78.84}  &{0.1360}      \\
{w/o TL}        & {0.83} & {0.29}  & {0.79} & {0.64} &{88.91}    & {0.0548}    \\
{Ours}&{\textbf{ 0.92}} & {\textbf{0.56}} & {\textbf{0.89}}  &{\textbf{ 0.79}} &{\textbf{94.22}}    &  {\textbf{0.0406}}  \\\hline
\end{tabular}
\label{tab:vm}
\end{table}
{ \noindent\textbf{Zero-shot objective evaluation.}
As no ground-truth instance matte is available in our data, we conduct a zero-shot testing on the public benchmark Video Matting~\cite{Erofeev2015} dataset. This dataset is built to evaluate the performance of the video matting results by given trimaps for each frame. Different from previous methods~\cite{zou2019unsupervised}, we focus on predicting smooth and accurate instance-level trimaps for video sequences.  Thus, we evaluate our method with several variants on the provided trimaps. The Intersection over Union (IoU) for the background (BG), trimap region (Tri) and foreground (FG) are reported. Besides, the smoothness is evaulated by the temporal consistency (TC) and the temporal stability (Tsb). All the metrics are the mean over all three types of given ground truth trimaps, including `narrow', `middle' and `wide'.
Note we only focus on portraits matting. Thus, only five out of ten video sequences from the original dataset are selected to build a person subset. The evaluated variants include: 1) SOLOV2~\cite{wang2020solov2} trained on the COCO dataset, which can only predict binary mask without trimap regions. 2) Our baseline, which is tuned on the Human Matting dataset, without training with the temporal loss and the pseudo labels on the video sequence.  3) Our baseline, which is trained on the combined dataset of Human Matting and the pseudo labels. 4) Our method, which is trained on the combined dataset with the temporal loss. The results are shown in Table~\ref{tab:vm}. We can see that learning from the video sequence can significantly improve the smoothness and the accuracy for the per-frame models. 
}

\noindent\textbf{Subjective
evaluation.}
{To better demonstrate the final results, we build four systems that can produce portraits. We %
use 
the IndexNet as the matting model for the four compared methods in the objective evaluation. The trimap of SOLOv2 is generated by the dilated and eroded operation. Other methods can directly predict trimap sequences. Results on ten video sequences by four methods are generated for the user study.}
We randomly show a video pair, including our methods and other methods in a random order, to the users. The users are required to mark a score from $1$ to $5$ to compare the quality of two video sequences ($5$ being `much better' and $1$ being `much worse'). Each video pair receives around $20$ scores. The final aggregated rating scores are shown in Table~\ref{tab:score}. We calculate the percentage of different opinions. Our methods significantly outperform compared methods showing that transfer learning methods on video sequence can improve the qualities of video portraits matting results.

\begin{table}[!h]
\vspace{-0.5em}
\setlength\tabcolsep{2pt}
	\centering
\small
		\caption{\small {Results of the user study on $10$ video sequences collected from YouTube. The 
		subjective 
		evaluation results
		demonstrate the effectiveness of the proposed training scheme boosts the performance of the per-frame models.} }
		\begin{tabular}{c|ccccc}
\hline
			 Ours vs. &  much better & better & similar & worse & much worse \\ 
\hline
			{ SOLO+Matt} & {23.6\%} & {52.7\%} & {18.9\%} & {4.7\%} & {0\%}\\
			{ w/o TL, PL} & {27.0\%} & {64.2\%} & {6.8\%} &{ 2.0\%} & {0\%}\\
			 w/o TL& 20.4\% & 61.1\% & 17.8\% & 0.1\% & 0\%\\
		\hline
		\end{tabular}

	\vspace{-1.5em}
	\label{tab:score}
\end{table}

\section{Conclusions}

In this work, we have proposed  to use per-frame models on video segmentation tasks to %
strike a balance between temporal consistency and 
accuracy. 
We design a few schemes to exploit the temporal consistency during training. 
The per-frame models are trained on multiple frames and processes  each frame separately during the inference.
The proposed training scheme is applied to real-time semantic video segmentation
and video instance segmentation.  %
On semantic video segmentation, by using compact networks and the new temporal consistency knowledge distillation, the proposed per-frame models outperform video-based methods significantly. Besides, our training scheme boosts the temporal consistency by a large margin for per-frame models.
{
On video instance segmentation, the proposed training scheme shows the potential of applying per-frame models to video sequences. For the application of portrait matting, the proposed training method shows %
promising results. 
}

{\small
\bibliographystyle{IEEEtran}
\bibliography{main}
}

\newpage

\vfill

\end{document}